\documentclass[journal]{IEEEtran}
\usepackage{graphicx}
\usepackage{booktabs} 
\usepackage{hyperref}
\usepackage{listings}
\usepackage{array}

\usepackage{algorithmicx}
\usepackage{algorithm}
\usepackage{algpseudocode}
\usepackage{algpseudocode}
\usepackage{amsmath}
\usepackage{multicol}
\usepackage{float}
\usepackage{amssymb}
\ifCLASSINFOpdf
\else
\fi
\begin{document}

\title{Exploring Natural Language-Based Strategies for Efficient Number Learning in Children through Reinforcement Learning}

\author{Tirthankar~Mittra,~Univeristy~Of~Colorado, Boulder}


\maketitle

\begin{abstract}
In this paper, we build a reinforcement learning framework to study how children compose numbers using base-ten blocks. Studying numerical cognition in toddlers offers a powerful window into the learning process itself, because numbers sit at the intersection of language, logic, perception, and culture. Specifically, we utilize state of the art (SOTA) reinforcement learning algorithms and neural network architectures to understand how variations in linguistic instructions can affect the learning process. Our results also show that instructions providing explicit action guidance are a more effective learning signal for RL agents to construct numbers. Furthermore, we identify an effective curriculum for ordering numerical-composition examples during training, resulting in faster convergence and improved generalization to unseen data. These findings highlight the role of language and multi-modal signals in numerical cognition and provide hypotheses for designing effective instructional strategies for early childhood education.
\end{abstract}


\IEEEpeerreviewmaketitle

\section{Introduction}
\IEEEPARstart{D}{eep} reinforcement learning has been successful in myriad kinds of tasks ranging from playing Atari games \cite{mnih2013playing}, to autonomous driving \cite{kiran2021deep} \cite{lin2025rl}, robotic manipulation~\cite{gu2016deep} \cite{kalashnikov2018qtopt}, recommendation system~\cite{10.1145/3543846} \cite{li2010contextual}, economics~\cite{sargent2018rl} \cite{moody2001learning} and even training large language models \cite{christiano2017deep} \cite{mroueh2025reinforcementlearningverifiablerewards} etc. There are also significant correspondences between reinforcement learning (RL) and the experimental study of animal learning in psychology. One notable example is how the temporal-difference (TD)~\cite{sutton1988learning} algorithm generalizes the Rescorla-Wagner model, whose main idea is that animals only learn when events violate their expectations \cite{walkenbach1980rescorla}. Another example is of reward shaping, a technique used in RL \cite{ng1999policy} to provide additional rewards to guide agents toward desired behaviors. This is an effective tool in both training RL agents and animals \cite{fugazza2015social}, similarly \cite{dayan2008decision} shows how RL explains human and animal learning processes, especially reward-based behavior. By systematically shaping the reward structure, trainers can accelerate the learning process. Furthermore, model based reinforcement learning, which involves the use of environmental models to predict future states and rewards, shares commonalities with what psychologists refer to as cognitive maps, these are mental representations that animals and humans use to navigate and understand their environment. Similarly, in RL the creation of an internal model of the environment allows agents to plan and make decisions based on predicted outcomes.\cite{tolman1948cognitive}.

Number learning in children is a crucial cognitive process that has been extensively studied both to understand artificial intelligence and to develop innovative pedagogical strategies \cite{monkevivciene2017pedagogical}. Our goal is to study this process by modeling it using reinforcement learning (RL), drawing inspiration from parallels that exist between RL and psychological learning theories. As per our knowledge, there haven’t been any efforts to model or gain insights into number composition in children using deep reinforcement learning techniques. In this paper we focus on the following three research questions. 
\textbf{(R1) Which input modalities are most critical for solving number composition tasks in RL agents?}
\textbf{(R2) What types of linguistic instructions most effectively support learning in these settings?}
\textbf{(R3) Does a structured training curriculum improve the efficiency and generalization of the number composition task?}

Additionally, we have developed a robust RL environment adhering to OpenAI API standards, which can be easily extended to further study various aspects of this cognitive learning process or other similar relational and hierarchical tasks. Using our reinforcement learning (RL) and language instruction framework, we discovered that agents learn significantly better when language instructions include explicit guidance on how to solve the task, compared to instructions that merely describe the state to the agent. Moreover, agents perform substantially worse and often fail to solve the task when only visual information is provided. Our model also uncovered an optimal ordering of numbers that consistently improves the agent's performance. Based on these findings, we propose that utilizing similar instructional strategies could greatly benefit children in their learning processes.

\section{Related Works}
Reinforcement learning (RL) agents face significant challenges when inferring abstract relational and causal structures within their environments, \cite{taylor2009transfer} discusses how standard RL struggles with relational abstractions and generalization. Recent studies have demonstrated that integrating language can significantly improve RL agent's understanding of environmental dynamics, thereby improving their ability to navigate and manipulate complex systems \cite{lampinen2022tell}
\cite{li2021implicit}. Moreover, language has been shown to assist RL agents in generalizing across different environments, facilitating the transfer of learned knowledge to new and varied contexts \cite{narasimhan2018grounding}. For humans, language explanations also play an important role in knowledge transfer and help us understand the causal structure of the world \cite{edmiston2015makes}. In this paper, we model learning of numbers in children using reinforcement learning and investigate how and which type of language instructions help RL agents acquire numerical concepts.

Prior work has extensively studied the development of numerical cognition in children. For instance, \cite{aunio2016core} uses longitudinal analysis to characterize the progression of core numerical skills in early childhood. However, such approaches do not incorporate machine learning or language-driven modeling frameworks. In this work, we address this gap by formulating number learning as a deep reinforcement learning problem. This formulation enables us to systematically analyze the factors that contribute to the acquisition of core numerical skills and to derive testable hypotheses regarding mechanisms that may accelerate learning. In particular, improvements observed in RL agents under specific training strategies may provide insights into effective instructional methods for children. Through this framework, we aim to advance the understanding of number learning mechanisms and contribute to the design of more effective educational interventions.

\section{Methodology}
\begin{figure}[!htbp]
    \centering
    \includegraphics[width=0.7\linewidth]{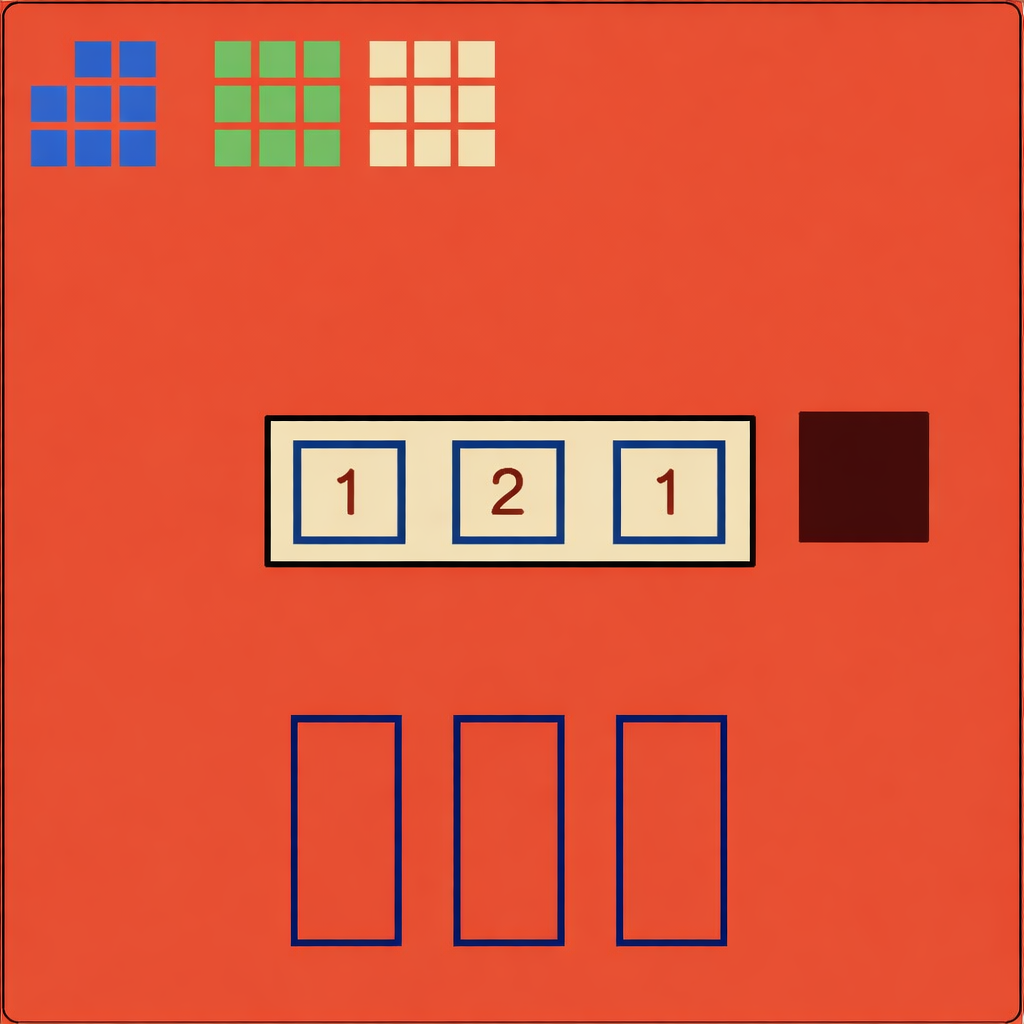}
    \caption{The image shows the visual information provided by the environment to the RL agent. The top right corner has blocks which are used to build the number displayed in the center.}
    \label{visualInfo}
\end{figure}

In this paper, we introduce a new reinforcement learning (RL) environment designed to study how language and visual cues affect children's ability to learn numbers. We model the child as an RL agent, and the agent's task is to build numbers using blocks that represent hundreds, tens, and ones. The agent has six possible actions. Three actions correspond to choosing blocks from the pool—one for hundreds, one for tens, and one for ones. The other three actions involve placing these blocks in the correct positions to ultimately form the number displayed by the environment. The agent receives visual information about the current state of the system (as shown in Figure [\ref{visualInfo}]) and uses different types of language instructions to complete the task.
If we observe the image, the number that the agent needs to construct is displayed at the center. In Figure [\ref{visualInfo}], the agent is tasked with constructing the number one hundred and twenty-one. The blue-colored box corresponds to the hundredth block, the green box corresponds to the tenth block, and the yellow block corresponds to the unit block. Beneath the number, there are three rectangles, each indicating the correct position where the agent should place the block. For instance, the agent must place one blue (hundredth) bock, two green (tenths) blocks and one yellow (unit) block below the digits one, two and one respectively to complete the task. For simplicity, an agent is only allowed to carry one type of block at a time. In Figure [\ref{visualInfo}], the agent has already picked up a hundredth block, which is represented by a black indicator box, on the right side of the Figure [\ref{visualInfo}]. If the agent is currently carrying any block, it will be represented by the black indicator box. This signal makes the task a Markov decision process for agents trying to solve it using only visual cues. Without this visual cue agents cannot distinguish between states where it is carrying a number block versus where it's not.

In our experimental setup, we provide our agent with two distinct types of language instructions to guide its decision. The first type, known as action based instructions, provides explicit directives on what actions the agent could take to accomplish the task at hand. These instructions essentially prescribe a strategy or policy that the agent can adhere to in order to navigate and solve the task efficiently. The second type of instructions, which we call state based instructions, offers descriptions of the current state of the environment to the agent. These descriptions convey the same information that the agent could otherwise gather by visually inspecting the environment. Any reinforcement learning (RL) task is fundamentally defined by three core components: the state, action, and reward, unlike the other components we don't provide reward-based language instructions because the environment is directly providing the reward to the agent, a linguistic description of the reward signal won't be valuable to solve the task.

\begin{figure}[!htbp]
    \centering
    \includegraphics[width=\linewidth]{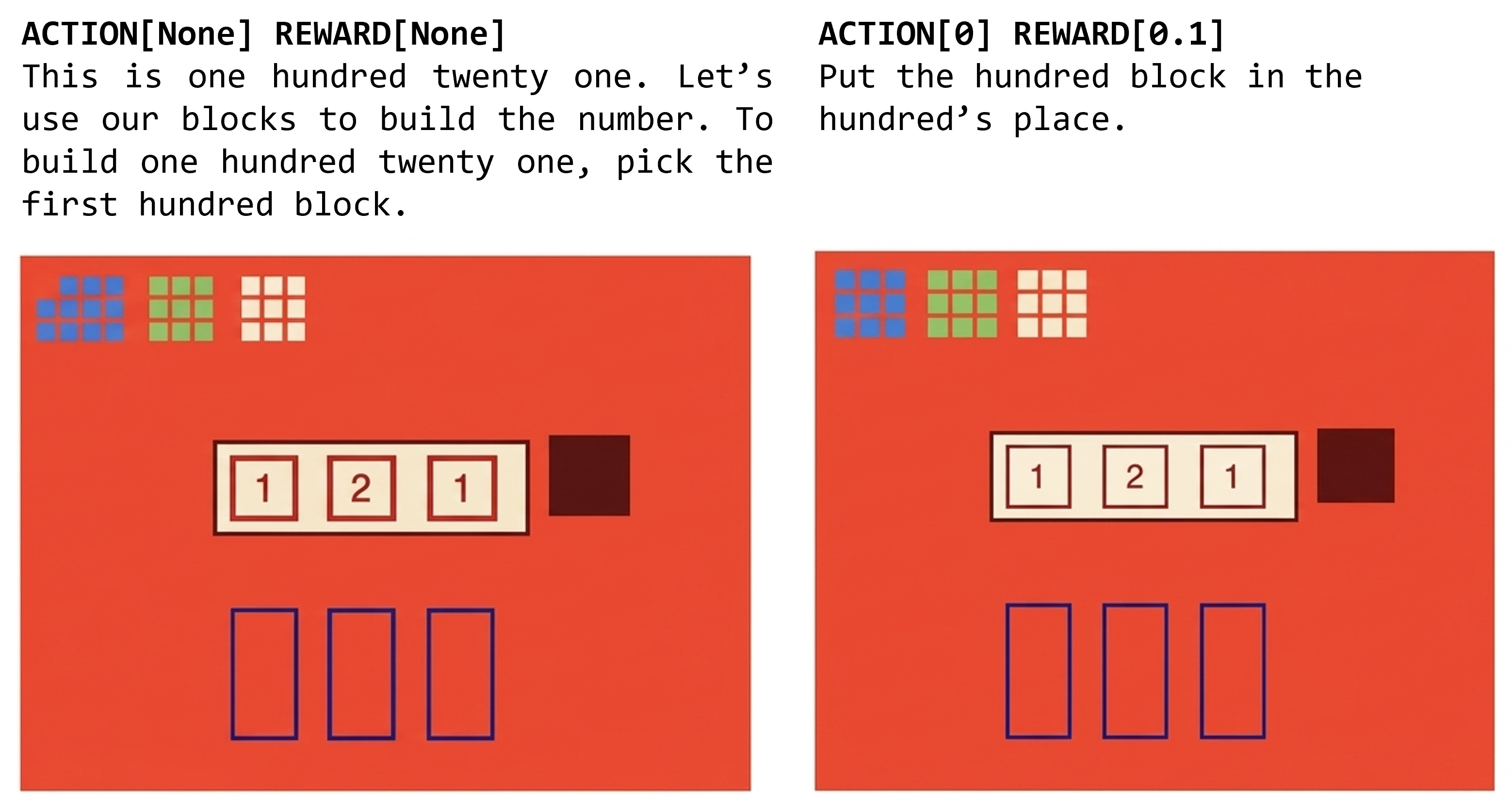}
    \caption{The image illustrates the state transition of an RL agent under a action-based instruction, with the previous state on the left, the next state on the right, and the language instruction shown at the top.}
    \label{policyBased}
\end{figure}

To understand how the environment operates, we will discuss some state transitions within our environment. The environment generates action based instructions as shown in Figure [\ref{policyBased}]. At the onset of each episode, the RL agent is provided with an initial image and an accompanying instruction. This instruction informs the agent of the displayed number and the action it should take to initiate solving the task. A proficiently trained agent will utilize both the visual information and the instruction to identify and pick up the hundredth block, transitioning to the subsequent state depicted in the second image of Figure [\ref{policyBased}]. However, it is not mandatory for the agent to strictly adhere to the actions suggested by the instruction. The detailed instructions provided to the agent are displayed in Figure[\ref{policyBased}].

\begin{figure}[!htbp]
    \centering
    \includegraphics[width=\linewidth]{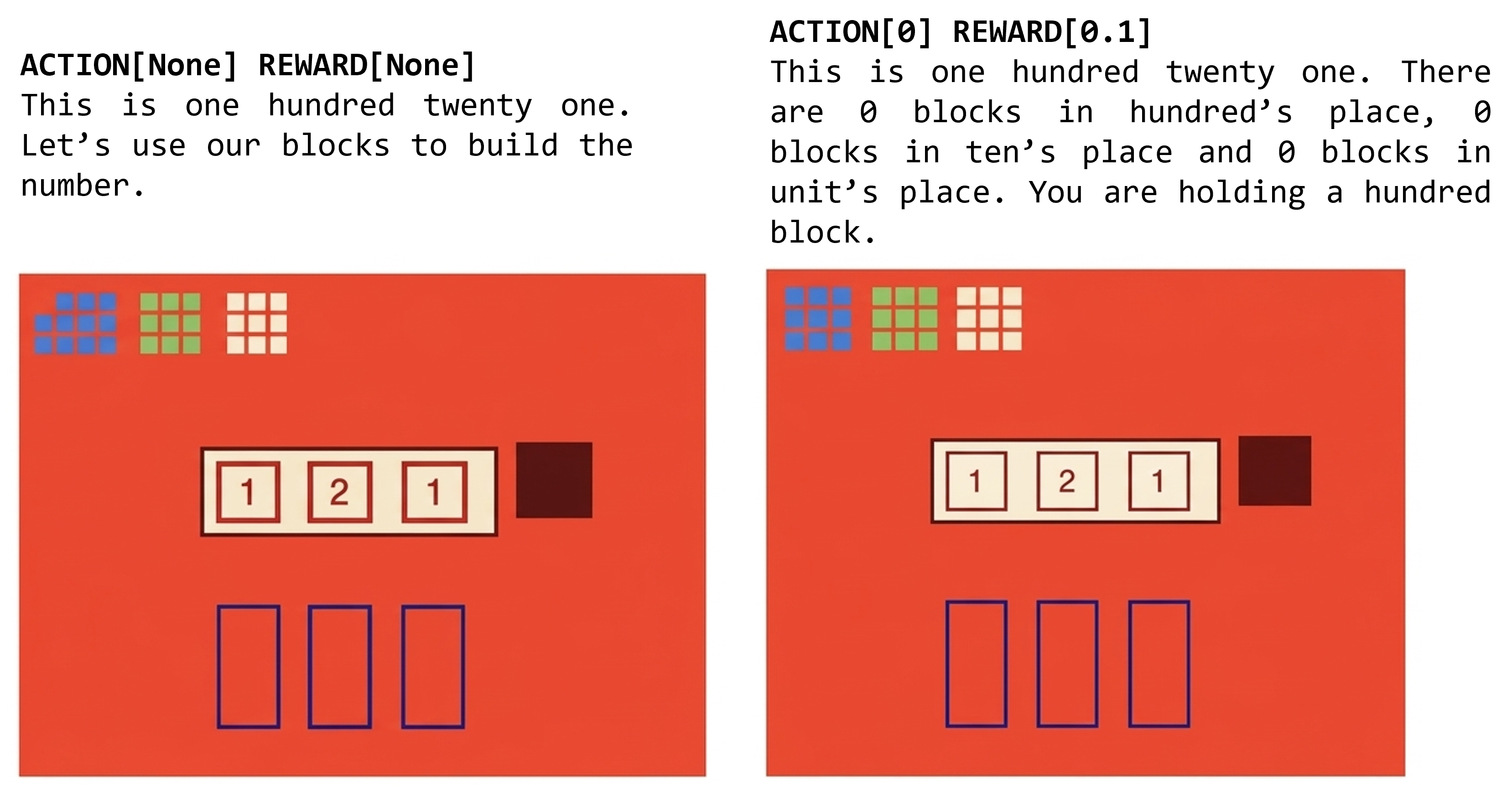}
    \caption{The image illustrates the state transition of an RL agent under a state-based instruction, with the previous state on the left, the next state on the right, and the language instruction shown at the top.}
    \label{stateBased}
\end{figure}

In state-based instructions, as illustrated in Figure [\ref{stateBased}], the environment provides a detailed description of the current state to the agent. For instance, in the example of forming the number one hundred and twenty-one, the initial instruction presents the number in words and prompts the agent to begin the task. A proficient agent, drawing from its experience, will select the correct block. In the subsequent instruction in the next state, the environment again informs the agent of the current number and that no blocks are currently placed in the hundredth, tenth, or unit place and that the agent is holding a hundredth block. One might question the necessity of repeating the first part of the instruction, "This is one hundred and twenty-one.". The repetition serves a crucial purpose, it provides the agent with an indirect signal of when to cease its actions and how many blocks it should still pick up, thereby making the task a Markov Decision Process(MDP), which is essential for using traditional reinforcement learning algorithms. The dynamics of an MDP adhere to the Markov property, which means that the system's future behavior depends solely on the present state and action. By reiterating the aforementioned instruction phrase, the agent is relieved of the burden of tracking past states, allowing it to make optimal decisions based solely on the current state. Agents exclusively utilizing language to solve the task needs to know the magnitude of each number it's trying to build in each state so it knows how many blocks to put for each digit and when to stop building the number.

Regarding the rewards received by the agent, two reward settings exist: sparse and dense. In the dense reward setting, the agent earns a reward of 0.1 for correctly placing a block in the designated position. Upon successfully forming the number by the episode's end, the agent receives a reward of +1. If the agent fails to complete the task within the allotted time or incorrectly places blocks, making the correct solution unattainable, a reward of -1 is given. The agent is allotted 2.5 times the optimal number of steps to solve the task. Exceeding this step count results in a -1 reward.
In the sparse reward setting, the agent receives either a +1 or -1 reward upon task completion or failure, respectively. A maximum step limit is set to prevent the agent from endlessly attempting to solve the task, all our experiments were conducted in the dense reward setting. In practical teaching scenarios, such as instructing a child to build numbers, constant feedback, whether through verbal instruction or rewards, is provided. Therefore, the dense reward setting more accurately reflects a real-life scenario and so we use this reward model for subsequent experiments.

We used some shortcuts to speed up our training process some of which can also be replicated in real life for teaching children. For example, when fine-tuning our reinforcement learning algorithm and deep neural network architecture, we kept an eye on the actions our agents took and the rewards they earned. If the rewards were lower than a certain threshold after a certain number of iterations, we discarded the model and made further adjustments. This helped us save a lot of time especially when doing hyper-parameter optimization(HPO). Additionally, we trained our model on numbers that could be formed using a small number of actions. Then, we let our trained model generalize to larger numbers. This helped speed up the training process. Specifically, we included all numbers between 1 and 999 in our training set that could be formed using no more than ten actions. The total number of episodes we used to train our model depended on the length of the training set. Each number in the set was given 500 episodes for the agent to learn. During training, the environment would display a number for 10 episodes and then move on to the next number in the training set. After showing all the numbers in the set, it would cycle back to the beginning. We found out that showing the same number for 500 episodes at a stretch produced worse results because the model got too fixated(over-fit) on one particular number and couldn't handle new numbers well, which made it harder for the model to learn and adapt to different situations. All experiments, fine-tuning, and development work were performed on a personal laptop with an Intel Core i7 processor and an Intel Iris Xe graphics card. The final training for all episodes was conducted on a virtual machine (VM) of the C2 series on the Google Cloud Platform. To prevent compilation and runtime errors resulting from differences in development and production environments, we created a Docker container which set up a consistent environment for our code to run.
 
\textbf{Experimental Setup:}
We conducted three broad experiments.
In the first experiment, we perform curriculum learning\cite{bengio2009curriculum}, where we put different numbers in the training set and test set based on various strategies to understand which strategy of building the training set provides the best result. For our subsequent experiments we use the strategy of building training set that yields the best result. Additionally, it would be interesting to see if children exhibit similar learning patterns to our agent when presented with the differently constructed training sets.

In our second experiment, we trained different Deep Neural Network Architectures to understand their relative performance in our RL environment. We took three different seeds and trained them independently to account for variance in results. After this experiment, we select the top-performing model as the one that most accurately mimics a child who is proficient at number learning.

The final experiment was to check which form of instruction provides better learning for our agent. Our reinforcement learning environment can have different constructs of instructions for example, the agent can be provided instructions in other languages. To assess the relative effectiveness of various forms of instruction, but to limit the scope of this paper we tried experimenting with two types of instructions. In one kind of instruction, we tell an agent what to do and in the other, we describe the state i.e. what is happening in the environment. The following subsections provide a detailed description of the key components of our proposed RL framework, including the reinforcement learning algorithm and the deep neural network architectures used.

The code is made publicly available at \href{https://github.com/tirthankar95/NumberLearningInChildren_RL_NLP}{code link}.

\subsection{Reinforcement Learning PPO}
In the reinforcement learning paradigm, agents interact with an environment to solve tasks and receive rewards based on their performance. Among the commonly used reinforcement learning algorithms we have deep Q-learning~\cite{mnih2013playing}, SARSA~\cite{rummery1994line}, vanilla policy gradient~\cite{sutton1999policy}, and actor-critic methods such as Trust Region Policy Optimization (TRPO)~\cite{schulman2015trust} and Proximal Policy Optimization (PPO)~\cite{schulman2017proximal}. Deep Q-learning, despite its popularity, suffers from several limitations: it is not well understood, often fails on simple tasks, and exhibits instability during training. Additionally, deep Q-learning requires a large number of training episodes to achieve satisfactory performance and involves tuning numerous hyper-parameters, including the learning rate, target network update frequency, and discount factor. A significant challenge with deep Q-learning is catastrophic interference, where new learning disrupts previously acquired knowledge~\cite{fedus2020catastrophicinterferenceatari2600}. Our preliminary experiments with DQN did not yield promising results, prompting us to explore policy gradient methods. However, vanilla policy gradient methods, such as REINFORCE~\cite{williams1992reinforce}, are known for high variance, which can impede learning. Consequently, we opted for the actor-critic method, specifically PPO, due to its ease of implementation and stable learning characteristics with comparatively lower variance. PPO is a model-free, off-policy learning algorithm that alternates between sampling data through interaction and optimizing a surrogate objective. Being an off-policy method, PPO is significantly more sample-efficient than traditional policy gradient methods like REINFORCE. The choice of PPO as our reinforcement learning algorithm due to its reduced number of hyper-parameters and enhanced stability also allowed us to concentrate on experimenting with deep neural network architectures rather than investing significant time in hyper parameter tuning. The most common policy gradient objective used for PPO is shown in Equation[\ref{eq1:PolicyGradientObjective}].
\begin{equation}
    J^{pg}(\theta) = \mathbb{E} \left[ A_t \cdot \frac{\pi_{\theta}(a_t \mid s_t)}{\pi_{\text{old}}(a_t \mid s_t)} \right] \label{eq1:PolicyGradientObjective}
\end{equation}
where, $A_{t}$ is called the advantage function, which quantifies how much better an action is compared to other actions in a given state $s_{t}$. The ratio $\frac{\pi_{\theta}(a_t \mid s_t)}{\pi_{\text{old}}(a_t \mid s_t)}$ is the importance sampling term used in typical off policy learning methods. Typical advantage estimates are calculated using the formula in Equation[\ref{eq2:AdvTypical}]
\begin{equation}
    A_{t} = r + \gamma * V(s_{t+1}) - V(s_{t})\label{eq2:AdvTypical}
\end{equation}
but we have used a generalized advantage estimate which reduces the variance of policy gradient estimates even more, Equation[\ref{eq3:GAE}].
\begin{equation}
    A_{t} = \delta_{t} + (\gamma \cdot \lambda) \cdot \delta_{t+1} + ... + (\gamma \cdot \lambda)^{T} \cdot \delta_{t+T} \label{eq3:GAE} 
\end{equation}
where, $$\delta_{t} = r_{t} + \gamma \cdot v_{\pi}(s_{t+1}) - v_{\pi}(s_
{t})$$
In the above Equation[\ref{eq3:GAE}], $(0 \leq \lambda \leq 1)$ is used to control bias and variance, higher value of $\lambda$ means higher variance. 
In our reinforcement learning framework, we incorporate an entropy term into the policy gradient objective (Equation[\ref{eq4:Entropy}]) to facilitate exploration within the environment. By maximizing entropy, the policy tends to select actions with lower probability, thereby promoting exploration, especially during initial training stages. This emphasis on maximizing entropy encourages the exploration of diverse actions, which is crucial for discovering optimal strategies in complex and uncertain environments. As training progresses, the policy gradually refines its focus based on accumulated experience, striking a balance between exploration and exploitation which achieves robust performance over time.
 \begin{equation}
     H(\pi(.|s)) = - \sum_{a} \pi(a|s) \cdot log(\pi(a|s))\label{eq4:Entropy}
 \end{equation}
Putting it all together, the final loss function, which we minimize using off-the-shelf optimizer, is shown below in Equation(\ref{eq5:Loss Function}).
\begin{equation}
\begin{aligned}
    L_{t} &= \alpha \cdot (r_{t} + v_{\pi}(s_{t+1}) - v_{\pi}(s_t))^{2} + \\
    &\quad - \beta \cdot \mathbb{E} \left[ A_t \cdot \frac{\pi_{\theta}(a_t \mid s_t)}{\pi_{\text{old}}(a_t \mid s_t)} \right] +\\
    &\quad - \gamma \cdot \sum_{a} \pi(a \mid s_{t}) \cdot \log(\pi(a \mid s_{t}))
\end{aligned}
\label{eq5:Loss Function}
\end{equation}
The first term in the Equation(\ref{eq5:Loss Function}) is the critic loss, the second term is the actor loss and the third term is the entropy term which encourages exploration.

In our study, we explore various neural network architectures to identify the architecture that most accurately replicates the learning process observed in children who are proficient at forming numbers. The goal here is to determine which neural network architecture achieves superior performance.

\subsection{Neural Network Architectures}
We evaluate four deep neural network models that learn the policy and value functions given a state ($s_t$). Each model is designed to study the role of different input modalities in the number composition task.

Detailed model architectures are provided in Appendix~\ref{ModelArch} but briefly, Model 1 uses only visual input, Model 2 uses both visual and linguistic inputs, Model 3 relies solely on linguistic input, and Model 4 also uses both modalities but combines them through an attention mechanism~\cite{vaswani2017attention} instead of fine-tuning ResNet and BERT like it's done in Model 2.

\section{Results}\label{Res}
The following section presents the outcomes of our experiments. We examined four different neural networks as described in the previous section. For all the results obtained from the reinforcement learning algorithm, we used the different seeds and plotted the 66\% confidence interval in the learning curves along with their mean, as illustrated in Figures [\ref{fig:pbi}] and [\ref{fig:sbi}]. Some of our experiments are used to validate our model by drawing parallels between children's learning processes, while others attempt to make predictions that can be applied to improve children's learning.

\subsection{Curriculum Learning}
During our experiments, we observed that the ordering of numbers in the training set significantly impacted the performance of the agents on both the training and test datasets. We explored three strategies for arranging data in the training set:

1. Ascending Order: Numbers were arranged in ascending order.

2. My Optimal Order: Numbers were sorted based on the ease with which an agent could solve the task. For example, an optimal agent exhausts the same number of actions to construct the number 1 and the number 100; for the number 1, the agent picks up and places the unit's block, and for the number 100, the agent does the same with the hundredth's block. Since these two numbers have the same ease of composition, they can be put as adjacent elements in the our order.

3. Descending Order: Numbers were arranged in descending order.

\begin{figure}[!htbp]
    \centering
    \includegraphics[width=0.5\textwidth]{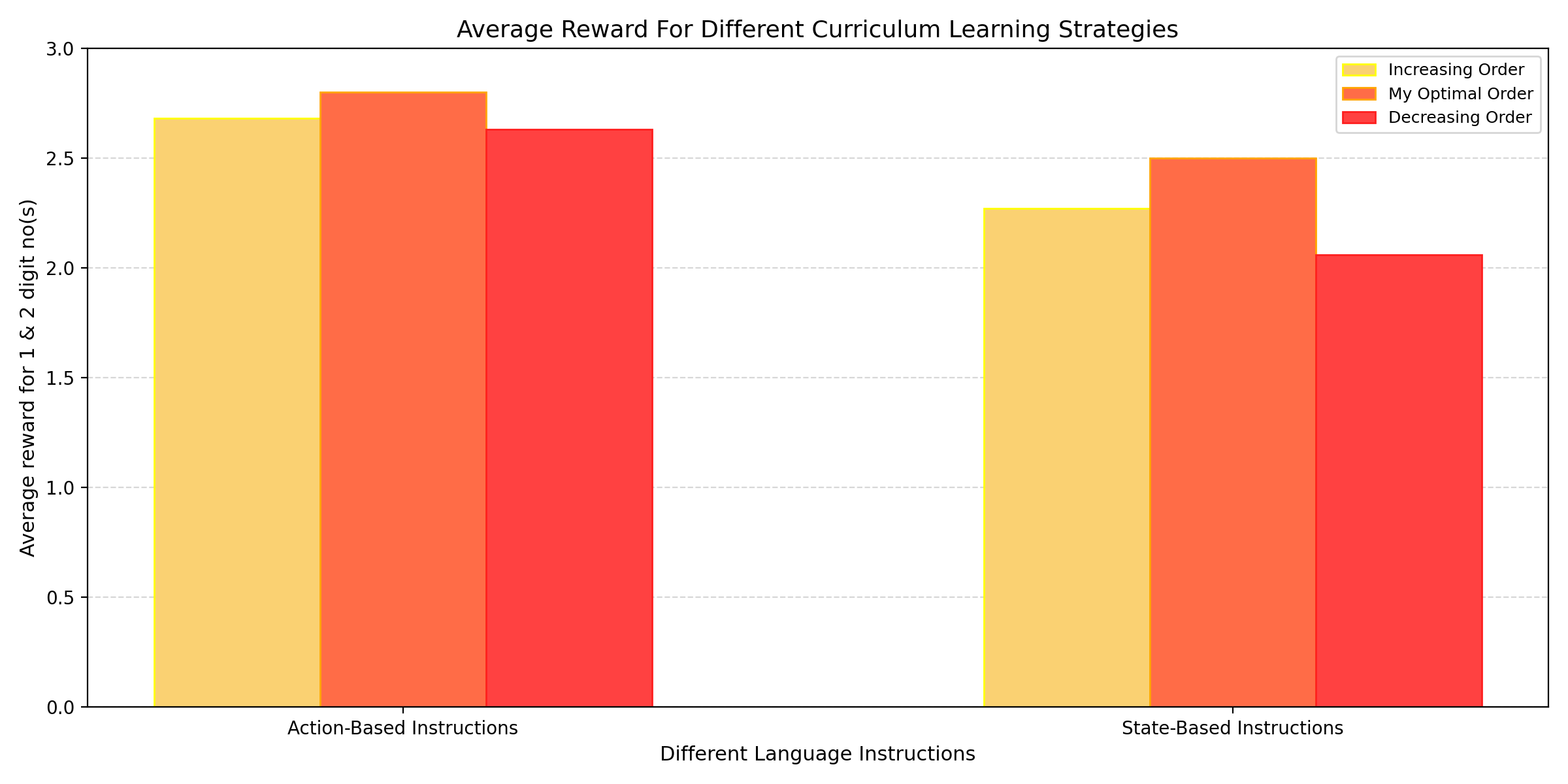}
    \caption{Average reward achieved by the attention-based RL agent under different curriculum learning strategies, evaluated separately for action-based (left) and state-based (right) instructions.}
    \label{fig:2D_CL_Tr}
\end{figure}

Our results indicated that the second strategy, where numbers were sorted by task-ease, consistently yielded the best performance across all models and different types of language instructions. Figure [\ref{fig:2D_CL_Tr}] illustrates the performance of the attention-based deep RL model(Model 4) with different instruction types for numbers between 0-99. The second strategy outperformed the first strategy (ascending order), followed by the third strategy (descending order). The poor performance of the third strategy can be attributed to the difficulty agents faced when required to solve larger numbers initially. Conversely, the first strategy allowed agents to solve numbers more easily at the beginning but ultimately performed worse than the second strategy. This was due to the development of a bias towards picking up the unit's block, as all numbers at the start of training can be solved using this strategy. This bias was detrimental when constructing numbers like 100, which do not contain unit digits. Given that agents performed better using the second type of ordering, and considering that we are modeling the learning of numbers, we can infer that human children might also benefit from being taught numbers in this order. Additionally, we experimented with training models using numbers arranged in a random order. The performance of the model with a random order fluctuated between that of the descending and task-ease orders. Due to this variance, we did not include the random strategy results in Figure [\ref{fig:2D_CL_Tr}], as consistent results are preferred over randomness.

\subsection{Performance On Action Based Instructions}
\begin{figure}[!htbp]
    \centering
    \includegraphics[width=0.5\textwidth]{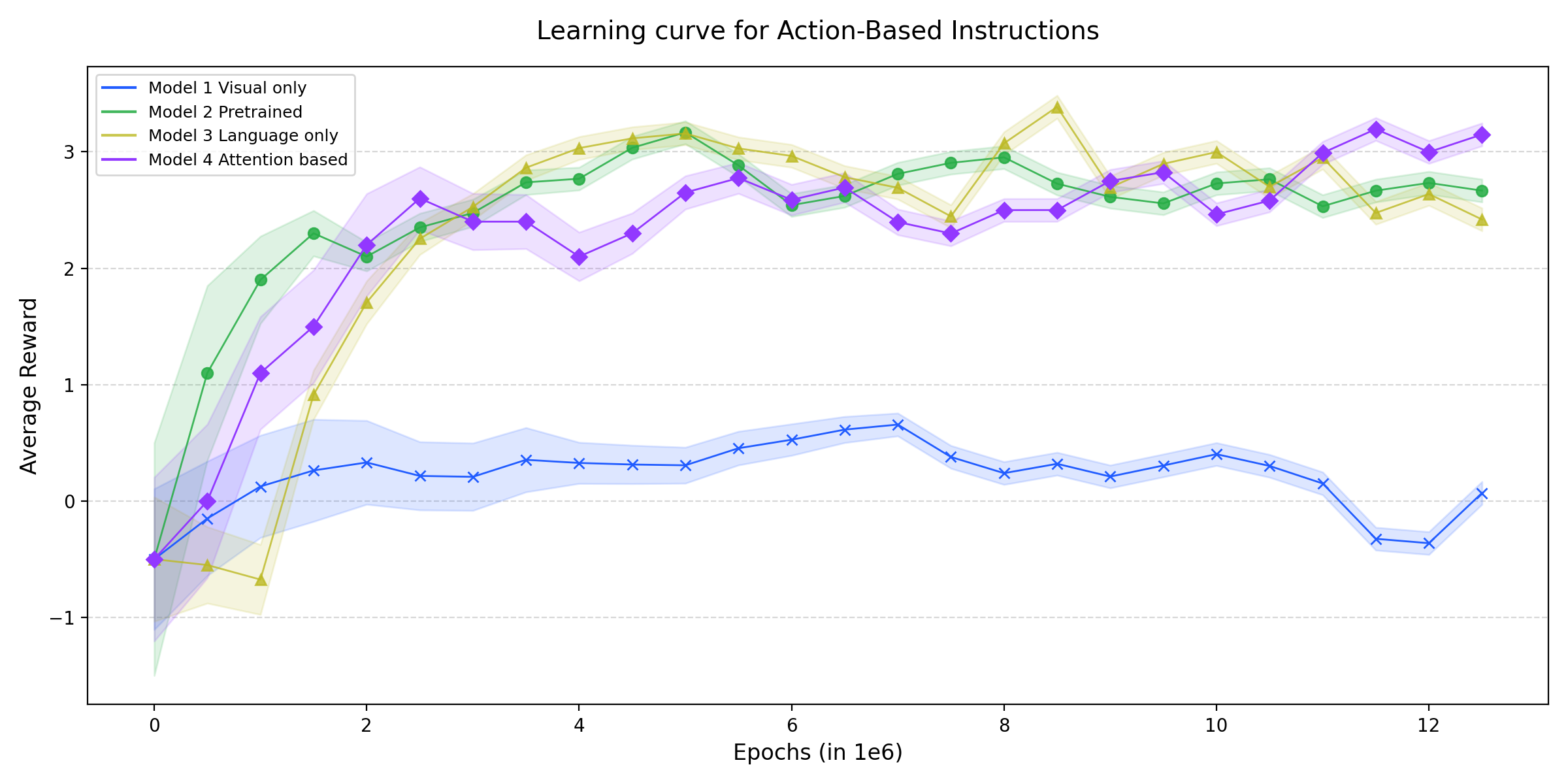}
    \caption{Illustration of learning curve when action-based instructions were used to train RL agents for various deep neural net architectures.}
    \label{fig:pbi}
 \end{figure}

\begin{figure}[!htbp]
    \centering
    \includegraphics[width=0.5\textwidth]{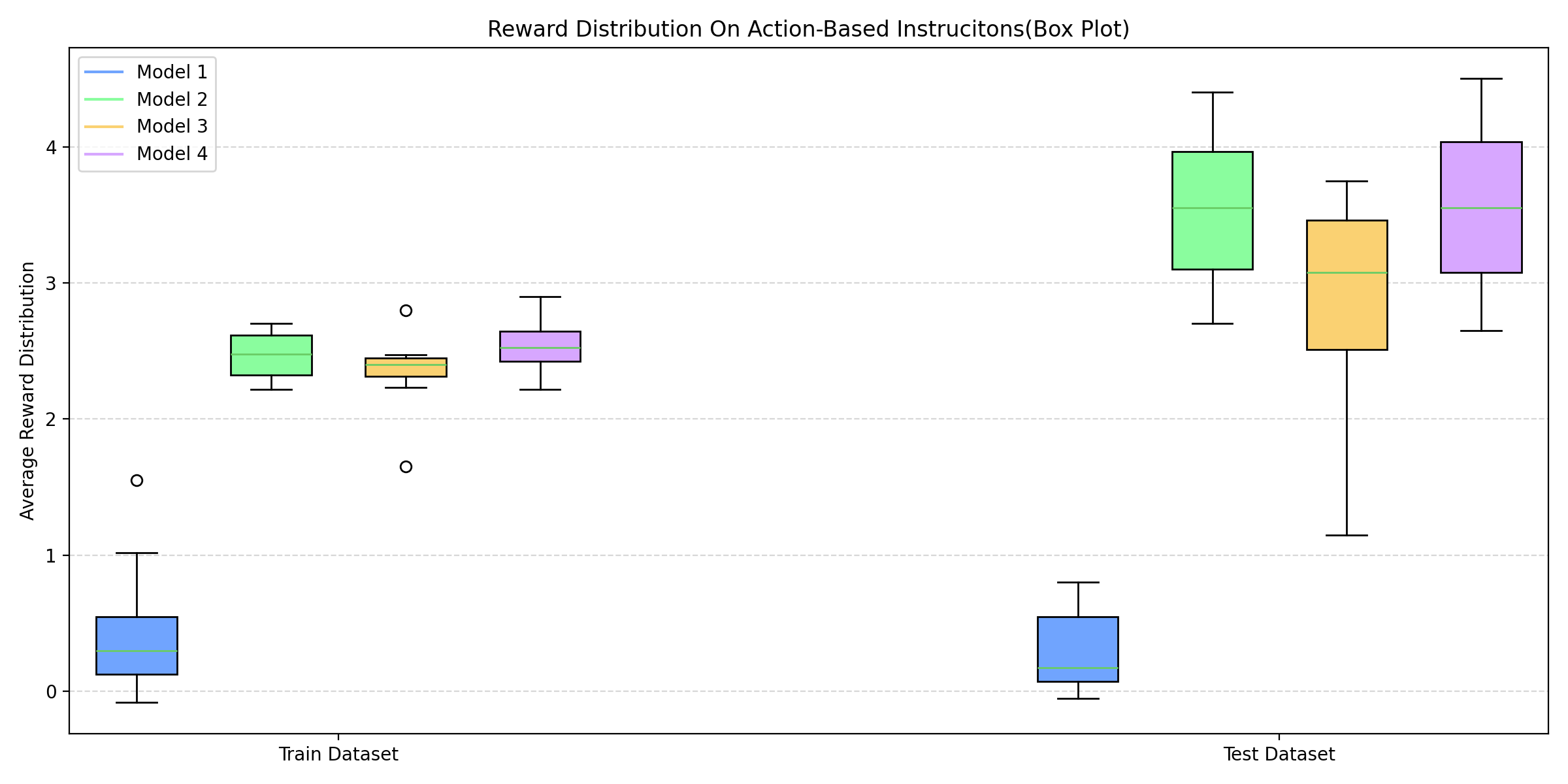}
    \caption{Image shows the distribution of average rewards(box-plot) for action-based instructions collected across various models for both training and test datasets.}
    \label{fig:pbi_te}
\end{figure}

In action-based instructions, the agent receives language instructions to guide its actions. Figure \ref{fig:pbi} displays the learning/training curves of our four neural network models. The learning curve plots the average reward on the entire training set periodically after a fixed number of frames, with each episode composed of multiple frames or states. The agent powered by an attention-based neural network(Model 4) achieved the highest average reward in the shortest number of frames. The worst-performing model was the agent that focused only on the visual information, ignoring the language instructions. After training, the performance of all models on the train and test dataset is shown in Figure \ref{fig:pbi_te}. The box plot illustrates the distribution of average rewards collected by the four models on both the training and test dataset. The average reward is lower in training data because it contains numbers which can be composed with lesser number of actions and hence by building numbers in training dataset the agent collects lower rewards compared the test data.

\subsection{Performance On State-Based Instructions}
\begin{figure}[!htbp]
    \centering
    \includegraphics[width=0.5\textwidth]{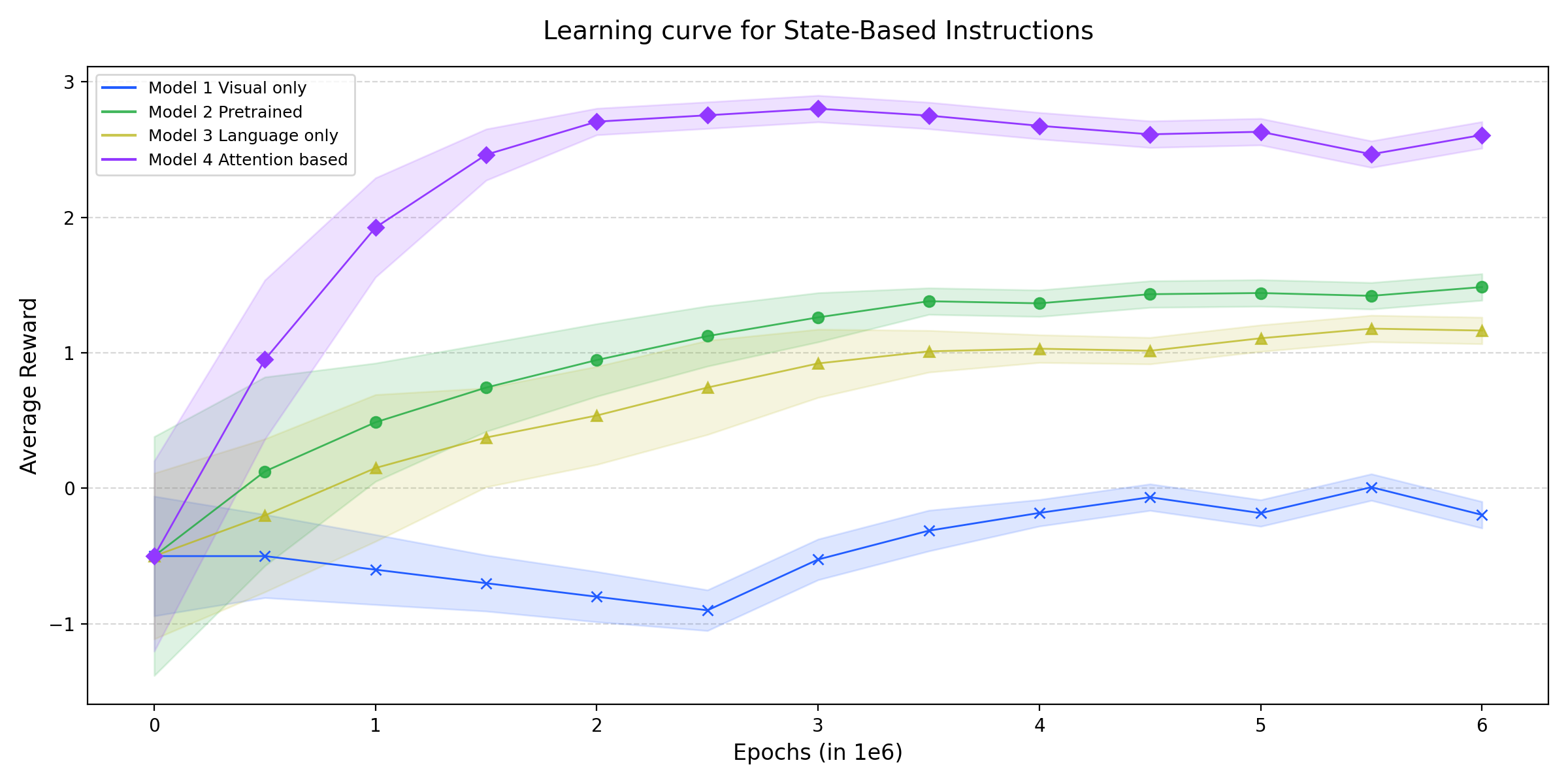}
    \caption{Illustration of learning curve when state-based instructions were used to train RL agents for various deep neural net architectures.}
    \label{fig:sbi}
\end{figure}

\begin{figure}[!htbp]
    \centering
    \includegraphics[width=0.5\textwidth]{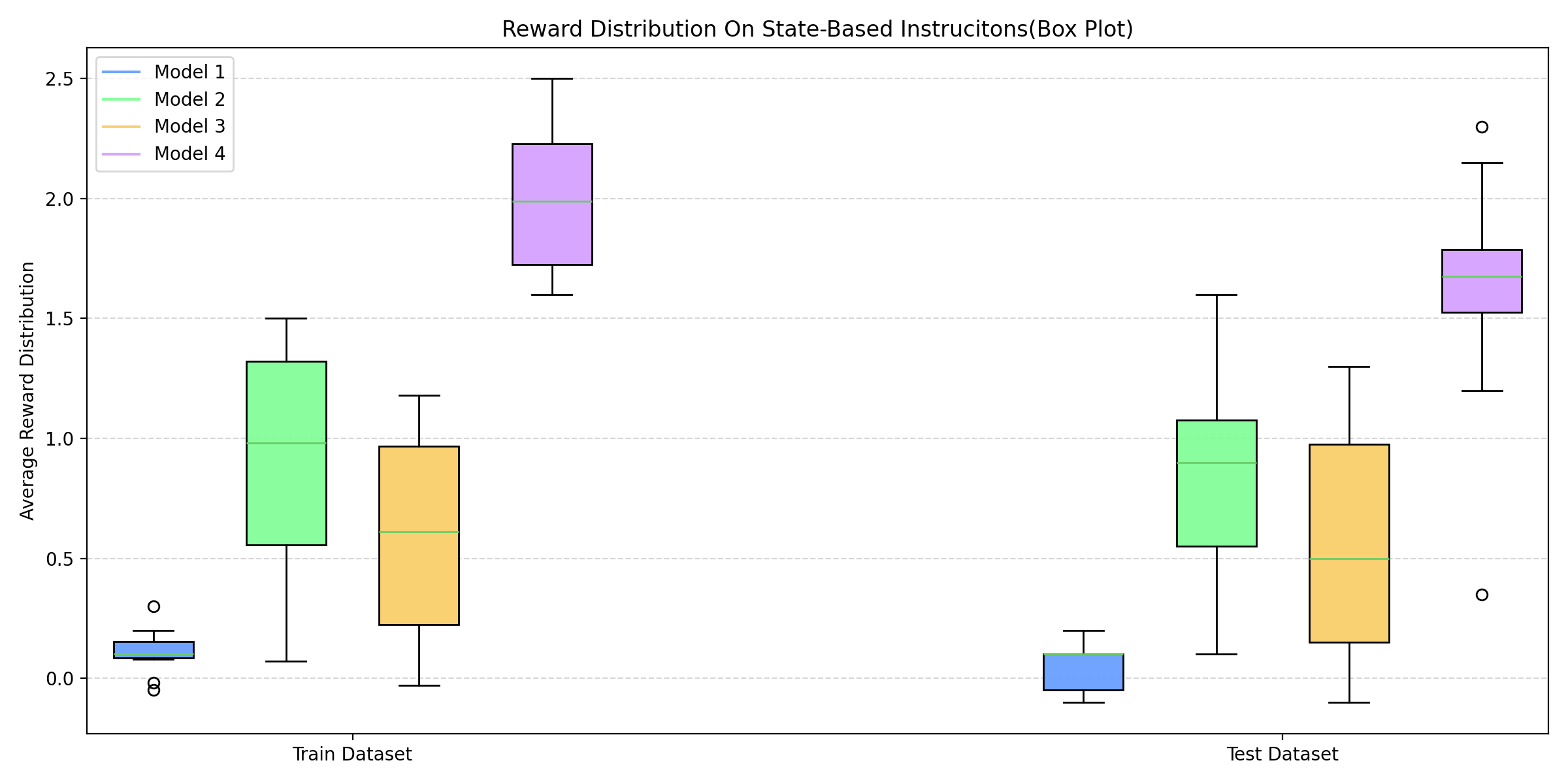}
    \caption{Image shows the distribution of average rewards(box-plot) for state-based instructions collected across various models for both training and test datasets.}
    \label{fig:sbi_te}
\end{figure}

In state based instructions, the agent is given language descriptions of the current state.  Figure[\ref{fig:sbi}] shows the learning curve. Average reward on the entire training set is calculated periodically after a certain number of frames and plotted on the learning curve. Figure[\ref{fig:pbi_te}] shows the performance of all four models on the training and test datasets, respectively. In general, for all the models performance on training data is better. Compared to results from action-based instructions, all models except the attention model fail to show good performance. In other words, the attention model stands out when the task's difficulty increases significantly. The attention model is also able to generalize well to numbers it has not seen, as can be seen in Figure[\ref{fig:sbi_te}]. Compared to the action-based instructions, agents faced more difficulty in solving the task. This performance gap can be attributed to the nature of state-based instructions, which describe the environment but do not specify the actions required to solve the task. As a result, the agent must infer the appropriate actions through interaction and experience, making the problem more challenging. Furthermore, the model that only uses visual information performs the worst. 

\textbf{Overall, from these two experiments, we can conclude that language instruction plays a more critical role for the agent to learn how to build numbers using base-10 blocks. Language can encode information more compactly compared to images for our task, and these linguistic instructions significantly speed up the learning process.}

A more granular breakdown of average rewards collected by the agents for different number ranges are provided in Appendix~\ref{Gran}

\subsection{Action based versus State based instructions}
\begin{figure}[!htbp]
    \centering
    \includegraphics[width=0.5\textwidth]{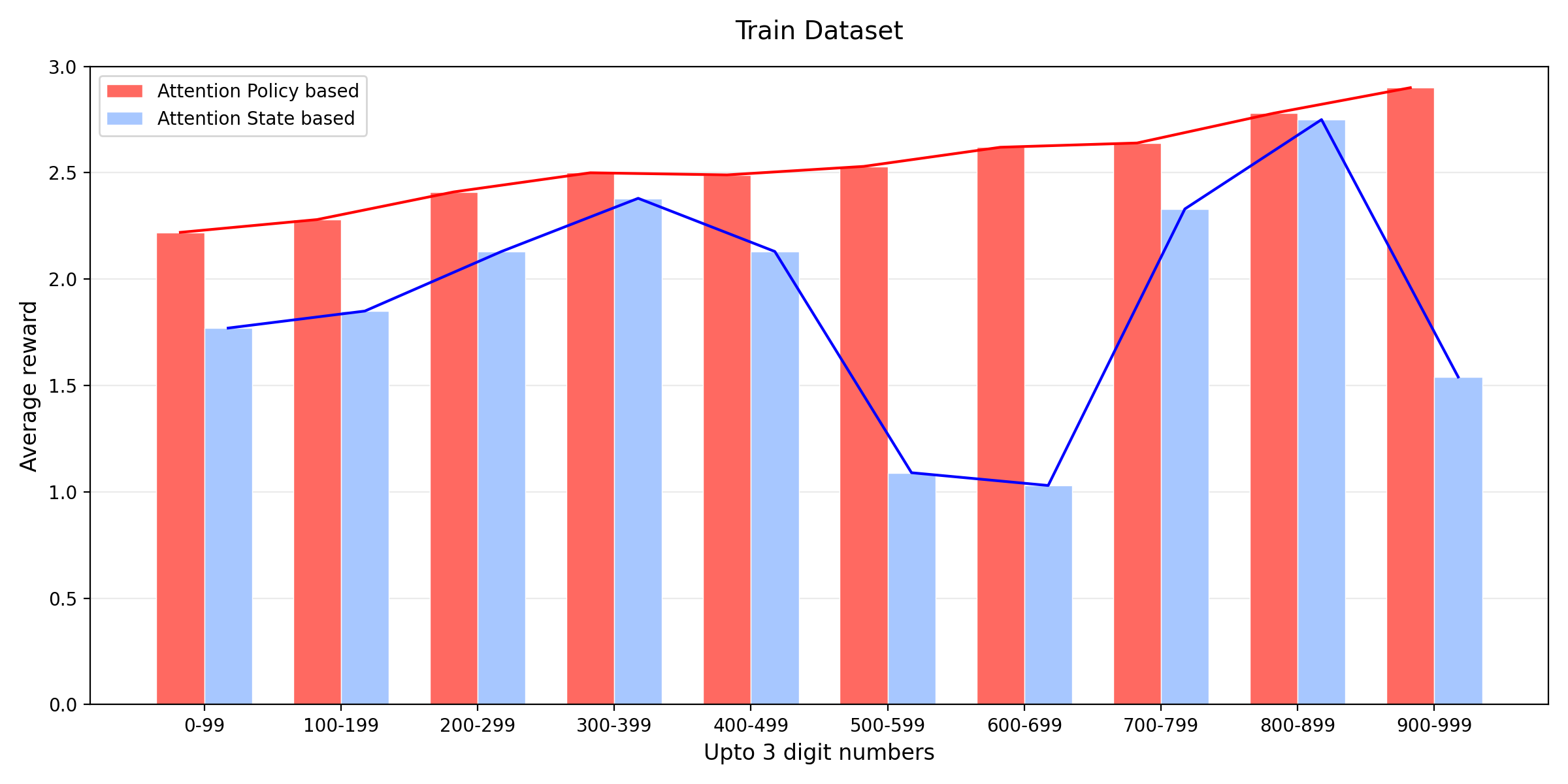}
    \caption{Comparison of the attention-based RL agent on action versus state-based instructions on the training dataset.}
    \label{fig:p_vs_s_tr}
\end{figure}

\begin{figure}[!htbp]
    \centering
    \includegraphics[width=0.5\textwidth]{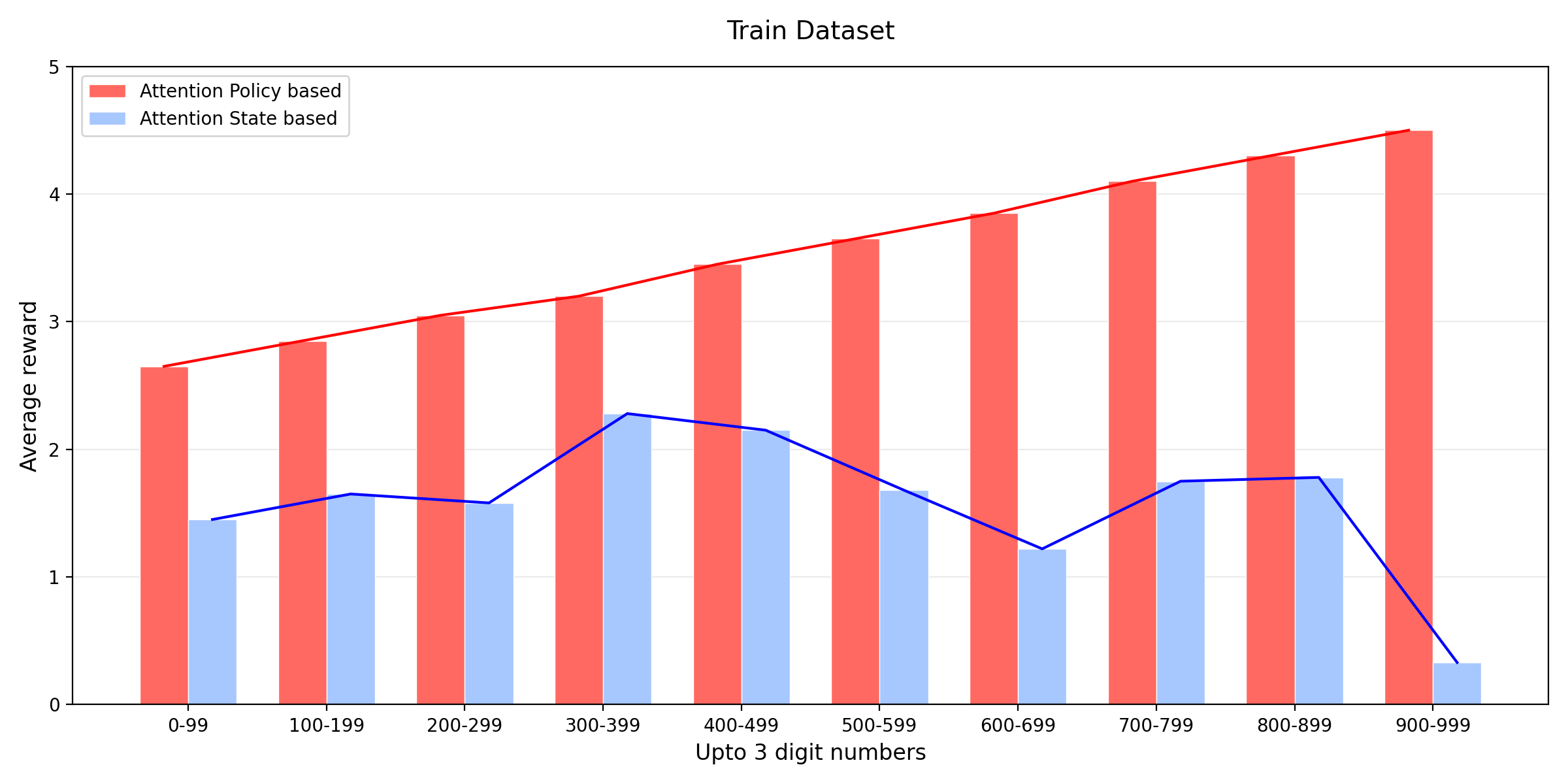}
    \caption{Comparison of the attention-based RL agent on action versus state-based instructions on the test dataset.}
    \label{fig:p_vs_s_te}
\end{figure}

Figures \ref{fig:p_vs_s_tr} and \ref{fig:p_vs_s_te} compares the performance of our best model i.e. the attention-based deep RL model, under two different types of instructions. This trend holds for all our models. The agent performs significantly better with action-based language instructions, and this difference is especially pronounced in the test set. The reason for this is that with action-based instructions, the agent doesn't need to figure out what actions to take for solving the task; it can simply follow the language instructions provided by the environment. In contrast, with state-based instructions, the agent doesn't know the optimal actions it needs to take. It must infer these actions by interacting with the reinforcement language environment. This setup is much more challenging for the agent to solve, resulting in the type of performance seen in Figures \ref{fig:p_vs_s_tr} and \ref{fig:p_vs_s_te}.

\section{Future Work}
One of our key contributions is a novel reinforcement learning environment, which can be utilized to explore how children grasp numerical concepts. With this innovative environment, our future goal is to investigate how different languages, such as Chinese, German, and Indian, among others, influence children's numerical learning. It's worth noting that we cannot utilize the pre-trained models for different languages due to the discrepancies in quality, based on the corpora they were trained on. To ensure equitable comparisons, we need to train models from scratch. Moreover, we intend to delve into the impact of various language structures. While our current study focuses on two types of instructions, namely state-based instructions that describe the environment and action-based instructions that suggest a policy for the agent, there are several other forms of language input that could be explored. For example, introducing noisy or irrelevant instructions alongside the true instructions could provide insights into the agent’s robustness and its ability to filter useful information.

\section{Conclusion}\label{conc}
In conclusion, we have observed numerous parallels between our model and the learning process of a child. For instance, if you instruct a child to construct numbers by simply following actions, they can easily comply. However, if you describe the situation or state using words and ask the child to solve the task, they may struggle to understand how to proceed. The same thing happens with our agent, if we give action based instruction to our agent meaning we are telling it what to do, it will have an easier time building the numbers but the agent doesn't learn anything about the underlying structure of the numbers or how to build them, it only learns to follow instructions blindly. Another parallel we observed was that when children are provided with more positive feedback, they learn faster compared to receiving sparse feedback. There was a similarity here with our model: when given dense rewards, it performed well, akin to how a child would perform under similar circumstances. We also observed that the order in which we arrange numbers in our training set has a different impact on the final model and its performance. It would be interesting to explore whether children also exhibit similar learning patterns. That is, if we teach them numbers in the order that yields the best performance for our model, would the children learn the numbers faster? In fact, we noticed that curriculum learning, where models learn more quickly with easier examples, is not the sole determining factor for optimal performance. It is also essential to ensure diversity among the easier examples to reduce model over-fitting. In our case, putting a proper mix of single-digit, two-digit, and three-digit numbers which can be formed easily in the training set resulted in better outcomes. Our experiments also revealed that language plays a more pivotal role in number composition task for children as our models couldn't solve the task only with visual cues. It seems for our task, language has the capacity to encode more information efficiently and in a more compact manner compared to visual information. Based on our initial two experiments, we can conclude that the attention-based model consistently outperforms other models, regardless of the type of linguistic instruction provided by the environment. 

\section{Acknowledgment}\label{ack}
This work was partially funded by DEL Lab and Prof. Lei Yuan at the University of Colorado Boulder. The authors would also like to acknowledge the computing resources provided by the lab and the University, which were instrumental in the completion of this research.

\ifCLASSOPTIONcaptionsoff
  \newpage
\fi

\bibliographystyle{IEEEtran}
\bibliography{ref}

@article{dayan2008decision,
  title={Decision theory, reinforcement learning, and the brain},
  author={Dayan, Peter and Daw, Nathaniel D},
  journal={Cognitive, affective, \& behavioral neuroscience},
  volume={8},
  number={4},
  pages={429--453},
  year={2008},
  publisher={Springer}
}

@misc{fedus2020catastrophicinterferenceatari2600,
      title={On Catastrophic Interference in Atari 2600 Games}, 
      author={William Fedus and Dibya Ghosh and John D. Martin and Marc G. Bellemare and Yoshua Bengio and Hugo Larochelle},
      year={2020},
      eprint={2002.12499},
      archivePrefix={arXiv},
      primaryClass={cs.LG},
      url={https://arxiv.org/abs/2002.12499}, 
}

@article{sutton1988learning,
  title={Learning to Predict by the Methods of Temporal Differences},
  author={Sutton, Richard S.},
  journal={Machine Learning},
  volume={3},
  number={1},
  pages={9--44},
  year={1988}
}

@inproceedings{li2010contextual,
  title={A Contextual-Bandit Approach to Personalized News Article Recommendation},
  author={Li, Lihong and Chu, Wei and Langford, John and Schapire, Robert E.},
  booktitle={WWW},
  year={2010}
}

@article{lin2025rl,
  title={Reinforcement Learning Methods for Autonomous Driving: A Survey},
  author={Lin, Yujie},
  journal={Applied and Computational Engineering},
  year={2025}
}

@article{kalashnikov2018qtopt,
  title={QT-Opt: Scalable Deep Reinforcement Learning for Vision-Based Robotic Manipulation},
  author={Kalashnikov, Dmitry and Irpan, Alex and Pastor, Peter and others},
  journal={arXiv preprint arXiv:1806.10293},
  year={2018}
}

@article{moody2001learning,
  title={Learning to Trade via Direct Reinforcement},
  author={Moody, John and Saffell, Matthew},
  journal={IEEE Transactions on Neural Networks},
  year={2001}
}

@book{sargent2018rl,
  title={Reinforcement Learning in Economics and Finance},
  author={Sargent, Thomas J. and Stachurski, John},
  year={2018},
  note={Lecture notes}
}

@inproceedings{vaswani2017attention,
  title={Attention Is All You Need},
  author={Vaswani, Ashish and Shazeer, Noam and Parmar, Niki and 
          Uszkoreit, Jakob and Jones, Llion and Gomez, Aidan N. and 
          Kaiser, {\L}ukasz and Polosukhin, Illia},
  booktitle={Advances in Neural Information Processing Systems (NeurIPS)},
  year={2017}
}

@article{williams1992reinforce,
  title={Simple statistical gradient-following algorithms for connectionist reinforcement learning},
  author={Williams, Ronald J.},
  journal={Machine Learning},
  volume={8},
  number={3-4},
  pages={229--256},
  year={1992}
}

@book{rummery1994line,
  title={On-line Q-learning using connectionist systems},
  author={Rummery, Gavin A and Niranjan, Mahesan},
  volume={37},
  year={1994},
  publisher={University of Cambridge, Department of Engineering Cambridge, UK}
}

@article{christiano2017deep,
  title={Deep reinforcement learning from human preferences},
  author={Christiano, Paul F and Leike, Jan and Brown, Tom and Martic, Miljan and Legg, Shane and Amodei, Dario},
  journal={Advances in neural information processing systems},
  volume={30},
  year={2017}
}

@misc{mroueh2025reinforcementlearningverifiablerewards,
      title={Reinforcement Learning with Verifiable Rewards: GRPO's Effective Loss, Dynamics, and Success Amplification}, 
      author={Youssef Mroueh},
      year={2025},
      eprint={2503.06639},
      archivePrefix={arXiv},
      primaryClass={cs.LG},
      url={https://arxiv.org/abs/2503.06639}, 
}

@article{10.1145/3543846,
author = {Afsar, M. Mehdi and Crump, Trafford and Far, Behrouz},
title = {Reinforcement Learning based Recommender Systems: A Survey},
year = {2022},
issue_date = {July 2023},
publisher = {Association for Computing Machinery},
address = {New York, NY, USA},
volume = {55},
number = {7},
issn = {0360-0300},
url = {https://doi.org/10.1145/3543846},
doi = {10.1145/3543846},
journal = {ACM Comput. Surv.},
month = dec,
articleno = {145},
numpages = {38}
}

@article{kiran2021deep,
  title={Deep reinforcement learning for autonomous driving: A survey},
  author={Kiran, B Ravi and Sobh, Ibrahim and Talpaert, Victor and Mannion, Patrick and Al Sallab, Ahmad A and Yogamani, Senthil and P{\'e}rez, Patrick},
  journal={IEEE transactions on intelligent transportation systems},
  volume={23},
  number={6},
  pages={4909--4926},
  year={2021},
  publisher={IEEE}
}

@article{mnih2013playing,
  title={Playing atari with deep reinforcement learning},
  author={Mnih, Volodymyr and Kavukcuoglu, Koray and Silver, David and Graves, Alex and Antonoglou, Ioannis and Wierstra, Daan and Riedmiller, Martin},
  journal={arXiv preprint arXiv:1312.5602},
  year={2013}
}

@article{edmiston2015makes,
  title={What makes words special? Words as unmotivated cues},
  author={Edmiston, Pierce and Lupyan, Gary},
  journal={Cognition},
  volume={143},
  pages={93--100},
  year={2015},
  publisher={Elsevier}
}

@inproceedings{he2016deep,
  title={Deep residual learning for image recognition},
  author={He, Kaiming and Zhang, Xiangyu and Ren, Shaoqing and Sun, Jian},
  booktitle={Proceedings of the IEEE conference on computer vision and pattern recognition},
  pages={770--778},
  year={2016}
}

@article{taylor2009transfer,
  title={Transfer learning for reinforcement learning domains: A survey.},
  author={Taylor, Matthew E and Stone, Peter},
  journal={Journal of Machine Learning Research},
  volume={10},
  number={7},
  year={2009}
}

@inproceedings{lampinen2022tell,
  title={Tell me why! Explanations support learning relational and causal structure},
  author={Lampinen, Andrew K and Roy, Nicholas and Dasgupta, Ishita and Chan, Stephanie CY and Tam, Allison and Mcclelland, James and Yan, Chen and Santoro, Adam and Rabinowitz, Neil C and Wang, Jane and others},
  booktitle={International Conference on Machine Learning},
  pages={11868--11890},
  year={2022},
  organization={PMLR}
}

@article{sutton1999policy,
  title={Policy gradient methods for reinforcement learning with function approximation},
  author={Sutton, Richard S and McAllester, David and Singh, Satinder and Mansour, Yishay},
  journal={Advances in neural information processing systems},
  volume={12},
  year={1999}
}

@article{schulman2017proximal,
  title={Proximal policy optimization algorithms},
  author={Schulman, John and Wolski, Filip and Dhariwal, Prafulla and Radford, Alec and Klimov, Oleg},
  journal={arXiv preprint arXiv:1707.06347},
  year={2017}
}

@inproceedings{schulman2015trust,
  title={Trust region policy optimization},
  author={Schulman, John and Levine, Sergey and Abbeel, Pieter and Jordan, Michael and Moritz, Philipp},
  booktitle={International conference on machine learning},
  pages={1889--1897},
  year={2015},
  organization={PMLR}
}

@article{li2021implicit,
  title={Implicit representations of meaning in neural language models},
  author={Li, Belinda Z and Nye, Maxwell and Andreas, Jacob},
  journal={arXiv preprint arXiv:2106.00737},
  year={2021}
}

@article{aunio2016core,
  title={Core numerical skills for learning mathematics in children aged five to eight years--a working model for educators},
  author={Aunio, Pirjo and R{\"a}s{\"a}nen, Pekka},
  journal={European early childhood education research journal},
  volume={24},
  number={5},
  pages={684--704},
  year={2016},
  publisher={Taylor \& Francis}
}

@article{narasimhan2018grounding,
  title={Grounding language for transfer in deep reinforcement learning},
  author={Narasimhan, Karthik and Barzilay, Regina and Jaakkola, Tommi},
  journal={Journal of Artificial Intelligence Research},
  volume={63},
  pages={849--874},
  year={2018}
}

@article{gu2016deep,
  title={Deep reinforcement learning for robotic manipulation},
  author={Gu, Shixiang and Holly, Ethan and Lillicrap, Timothy P and Levine, Sergey},
  journal={arXiv preprint arXiv:1610.00633},
  volume={1},
  pages={1},
  year={2016}
}

@article{tolman1948cognitive,
  title={Cognitive maps in rats and men.},
  author={Tolman, Edward C},
  journal={Psychological review},
  volume={55},
  number={4},
  pages={189},
  year={1948},
  publisher={American Psychological Association}
}

@inproceedings{ng1999policy,
  title={Policy invariance under reward transformations: Theory and application to reward shaping},
  author={Ng, Andrew Y and Harada, Daishi and Russell, Stuart},
  booktitle={Icml},
  volume={99},
  pages={278--287},
  year={1999}
}

@article{walkenbach1980rescorla,
  title={The Rescorla-Wagner theory of conditioning: A review of the literature},
  author={Walkenbach, John and Haddad, Nabil F},
  journal={The Psychological Record},
  volume={30},
  pages={497--509},
  year={1980},
  publisher={Springer}
}

@article{fugazza2015social,
  title={Social learning in dog training: The effectiveness of the Do as I do method compared to shaping/clicker training},
  author={Fugazza, Claudia and Mikl{\'o}si, {\'A}d{\'a}m},
  journal={Applied Animal Behaviour Science},
  volume={171},
  pages={146--151},
  year={2015},
  publisher={Elsevier}
}

@inproceedings{monkevivciene2017pedagogical,
  title={Pedagogical strategies that improve children’s play-based learning},
  author={Monkevi{\v{c}}ien{\.e}, Ona and Stankevi{\v{c}}ien{\.e}, Kristina and Autukevi{\v{c}}ien{\.e}, Birut{\.e} and Jonilien{\.e}, Marija},
  booktitle={SOCIETY. INTEGRATION. EDUCATION. Proceedings of the International Scientific Conference},
  volume={2},
  pages={290--307},
  year={2017}
}

@article{devlin2018bert,
  title={Bert: Pre-training of deep bidirectional transformers for language understanding},
  author={Devlin, Jacob and Chang, Ming-Wei and Lee, Kenton and Toutanova, Kristina},
  journal={arXiv preprint arXiv:1810.04805},
  year={2018}
}

@inproceedings{bengio2009curriculum,
  title={Curriculum learning},
  author={Bengio, Yoshua and Louradour, J{\'e}r{\^o}me and Collobert, Ronan and Weston, Jason},
  booktitle={Proceedings of the 26th annual international conference on machine learning},
  pages={41--48},
  year={2009}
}

\begin{IEEEbiographynophoto}{Tirthankar Mittra} received his B.E. degree in electronics and telecommunication engineering from Jadavpur University, Kolkata, India, and his M.S. degree in computer science from the University of Colorado Boulder, Boulder, CO, USA. He has industry experience at Samsung R\&D, where he contributed to multiple patent filings and peer-reviewed research publications. He has also worked at Qualcomm and Grid Dynamics as a software engineer/machine learning engineer/data scientist. His current research interests include reinforcement learning(multi-agent), natural language processing, and the application of AI methods to automated problem generation and reasoning systems.
\end{IEEEbiographynophoto}

\clearpage

\appendices
\section{Neural Network Architectures Powering Reinforcement Learning, PPO}\label{ModelArch}
\subsubsection{Model 1}
\begin{figure}[!htbp]
	\centering
	\includegraphics[width=0.35\textwidth]{ 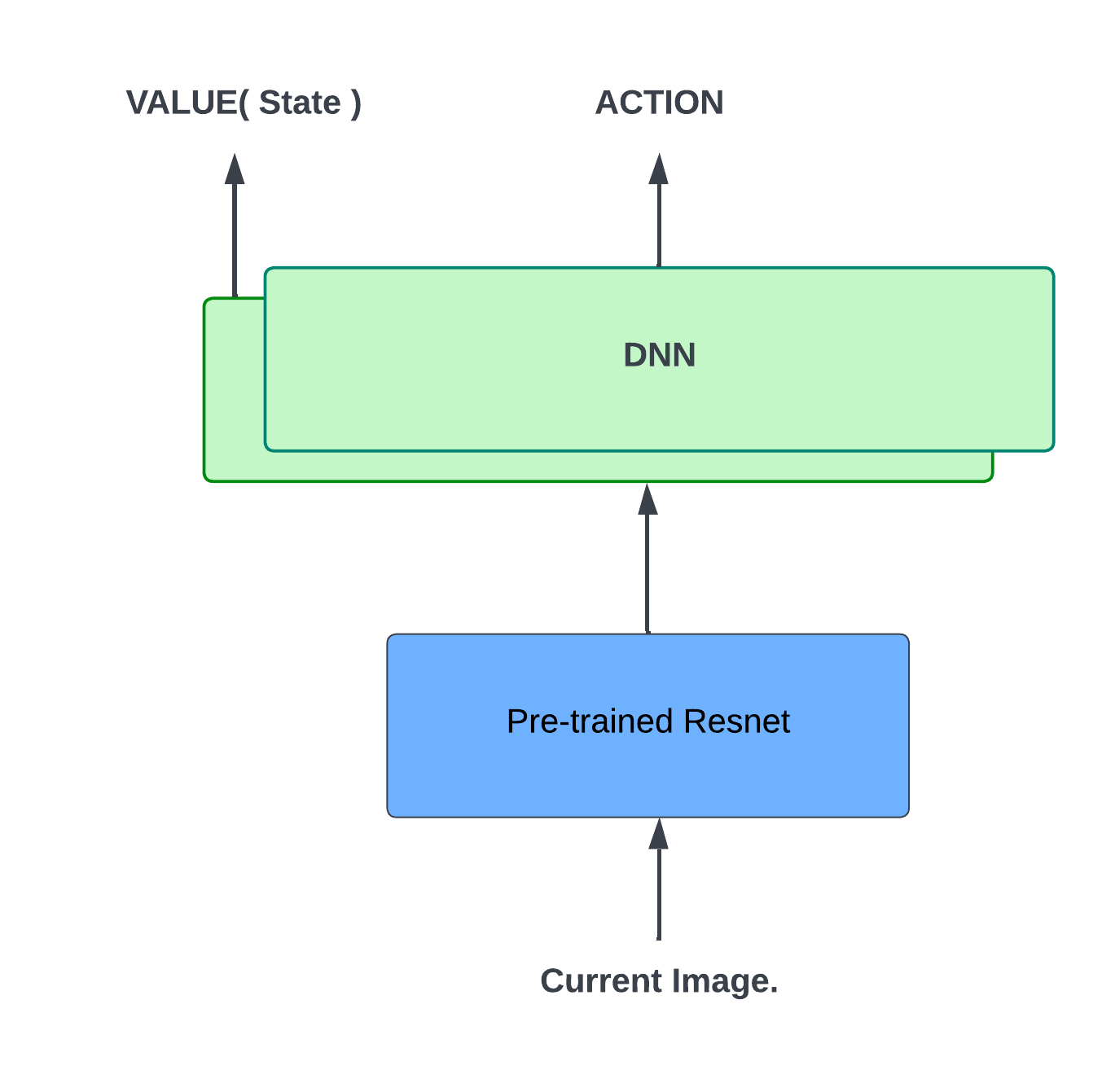}
	\caption{Model 1: Fine-tuned ResNet which relies on only visual information to solve the task of building numbers.}
	\label{Model1}
\end{figure}

In this model, the agent relies exclusively on visual observations from the environment to learn and find the optimal policy. The agent receives raw image frames as input and must extract all relevant information from these images alone, without any additional language or semantic guidance.
Figure~\ref{Model1} illustrates the neural network architecture used for this model. As the backbone, we use a pre-trained ResNet~\cite{he2016deep}, a deep convolutional neural network originally trained on large-scale image recognition tasks. By leveraging a pre-trained ResNet, the agent benefits from rich visual features already learned from millions of images, rather than learning visual representations from scratch. To allow the model to adapt to our specific task, we unfreeze and make the top layers of the ResNet trainable, enabling the network to fine-tune its feature extraction for the reinforcement learning environment.
The output of the ResNet backbone is then passed into two separate dense neural network heads, each serving a distinct purpose. The first dense head acts as the \textit{value network}, outputting an estimate of the state value $v_{\pi}(s_t)$ — a scalar representing how beneficial it is for the agent to be in the current state $s_t$ under policy $\pi$. The second dense head acts as the \textit{policy network}, predicting the action the agent should take given the current visual observation. This two-headed architecture falls under the actor-critic method, where the value estimate helps stabilize and guide the policy update during training.

\begin{figure}[!htbp]
	\centering
	\includegraphics[width=0.4\textwidth]{ 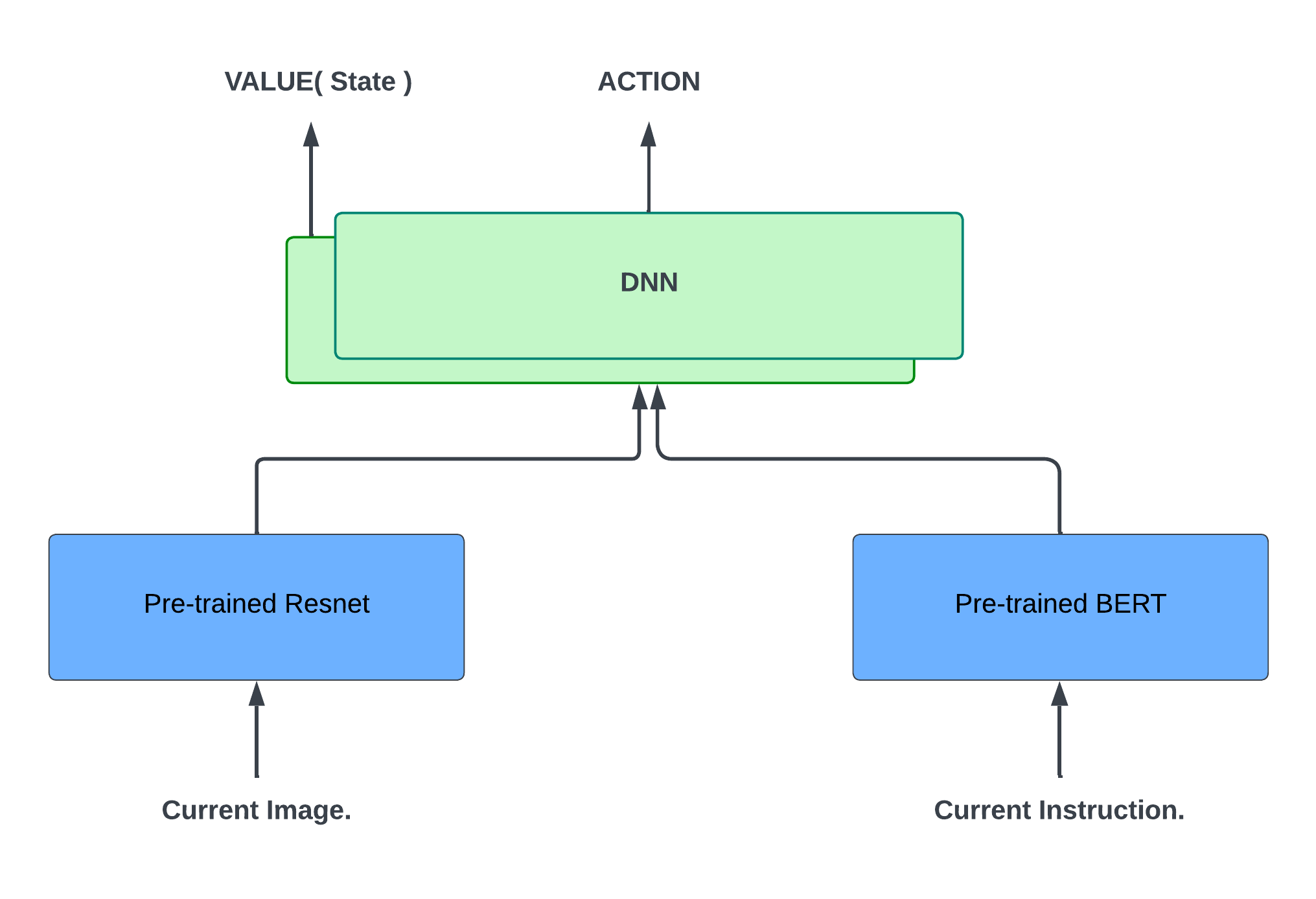}
	\caption{Model 2: A Pre-trained BERT and Fine-tuned ResNet model which utilizes both modalities to solve the number composition task.}
	\label{Model2}
\end{figure}

\subsubsection{Model 2}
The agent having model 2 as it's neural network architecture uses both visual and language instruction to make a decision. This model uses pre-trained Neural Network architecture BERT\cite{devlin2018bert} and ResNett~\cite{he2016deep}. To adapt these models to our task, we remove the original classification head and unfreeze the top layers of ResNet, allowing it to learn task-specific features. The ordering of these components, by keeping them at the bottom, ensures that the agent utilizes the representations from both BERT and ResNet. The idea to use these pre-trained models was to find out if there is a knowledge transfer from tasks on which these Neural Networks were trained on.

\subsubsection{Model 3}
In the third neural network model, we use a simple fully connected (dense) architecture without any attention mechanisms or pre-trained components. This design choice allows us to isolate and evaluate the contribution of more complex architectures, such as attention layers and pre-trained models, to the agent’s learning process. By removing these components, Model 3 serves as a baseline to understand whether simpler architectures are sufficient to solve the task. To ensure a fair comparison with Model 2, we keep the number of trainable parameters in Model 3 approximately the same. This helps control for model capacity, so that any differences in performance can be attributed primarily to architectural choices rather than differences in the number of parameters. This model only uses linguistic signals to solve the task. 

\begin{figure}[!htbp]
	\centering
	\includegraphics[width=0.4\textwidth]{ 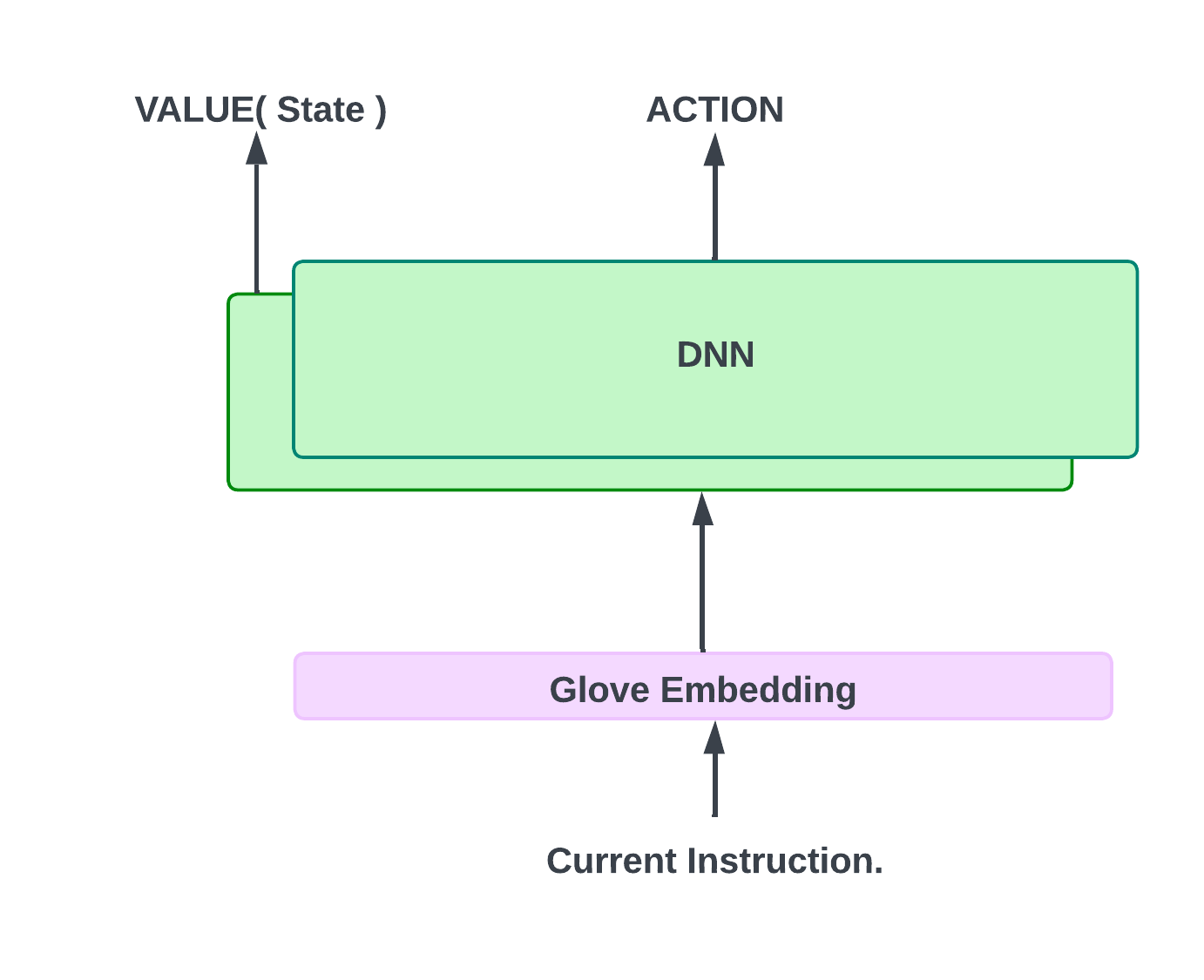}
	\caption{Model 3: A lightweight dense neural network model which only utilizes language instructions to solve number composition task.}
	\label{Model3}
\end{figure}

\begin{figure}[!htbp]
	\centering
	\includegraphics[width=0.5\textwidth]{ 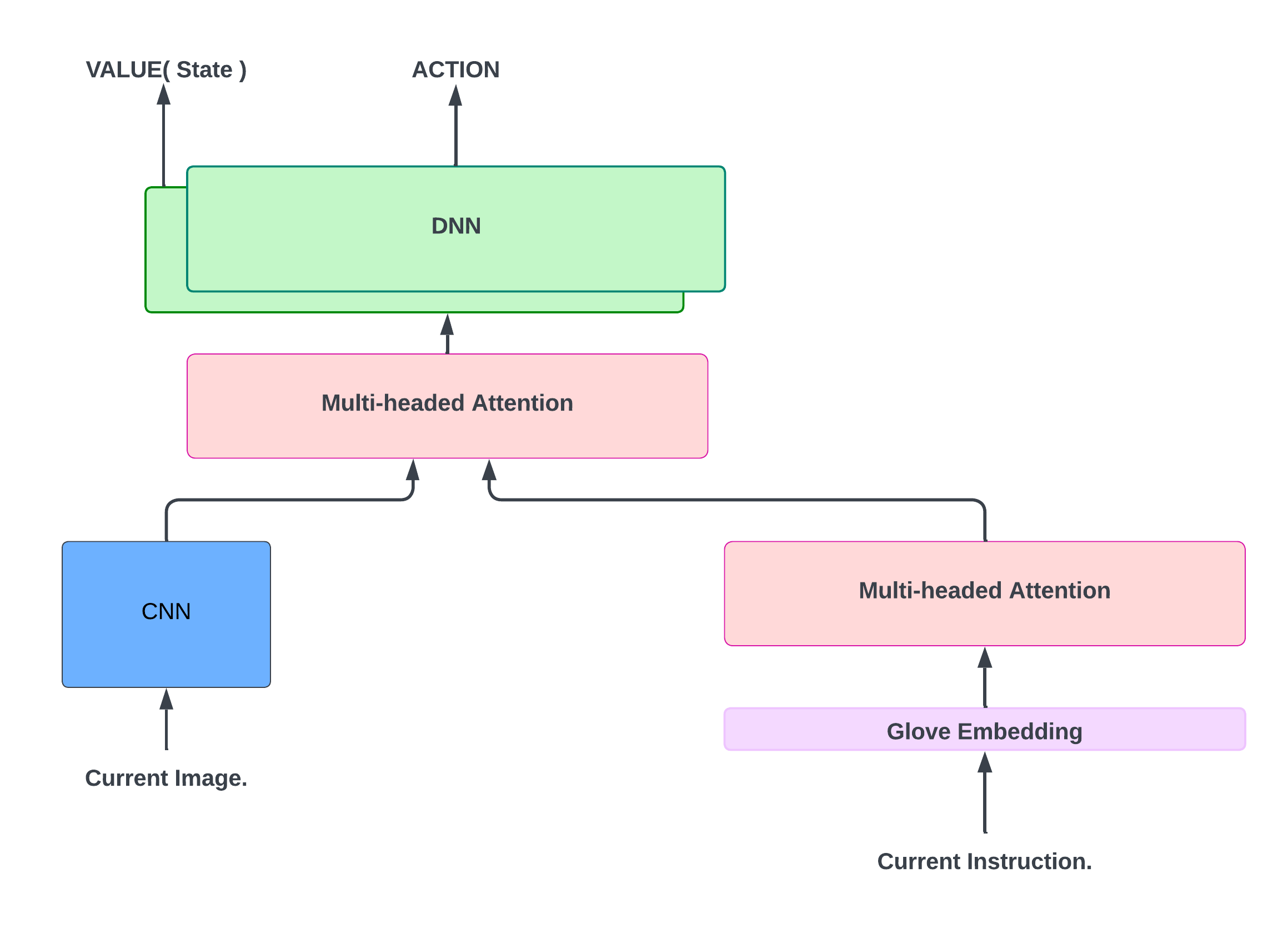}
	\caption{Model 4: An attention-based model trained from the scratch that mixes both visual and language information using an attention layer to solve the number building task.}
	\label{Model4}
\end{figure}

\subsubsection{Model 4}
In the final model, we incorporate an attention layer that is trained from scratch. Unlike Model 2, which relies on pre-trained architectures, this model is designed to learn task-specific representations entirely during training. Training large-scale models such as BERT or ResNet from scratch is computationally expensive and time-consuming. To address this, we design a smaller, custom attention-based model with reduced complexity.This allows us to study the effectiveness of attention mechanisms in isolation, while keeping the computational requirements manageable. By comparing this model with others, we can better understand whether attention alone, without pre-training, provides meaningful benefits for the task.

\section{Granular Average Rewards Across Different Language Instructions And Neural Network Architectures} \label{Gran}

The bar plots below illustrates the average rewards collected by the four deep RL models across various number ranges. The results are shown separately for different types of language instructions across training and test datasets. Note, that the maximum possible average reward increases as the number ranges on the x-axis increase because if an agent builds a bigger number, it receives a higher reward.

\begin{figure}[!htbp]
	\centering
	\includegraphics[width=0.5\textwidth]{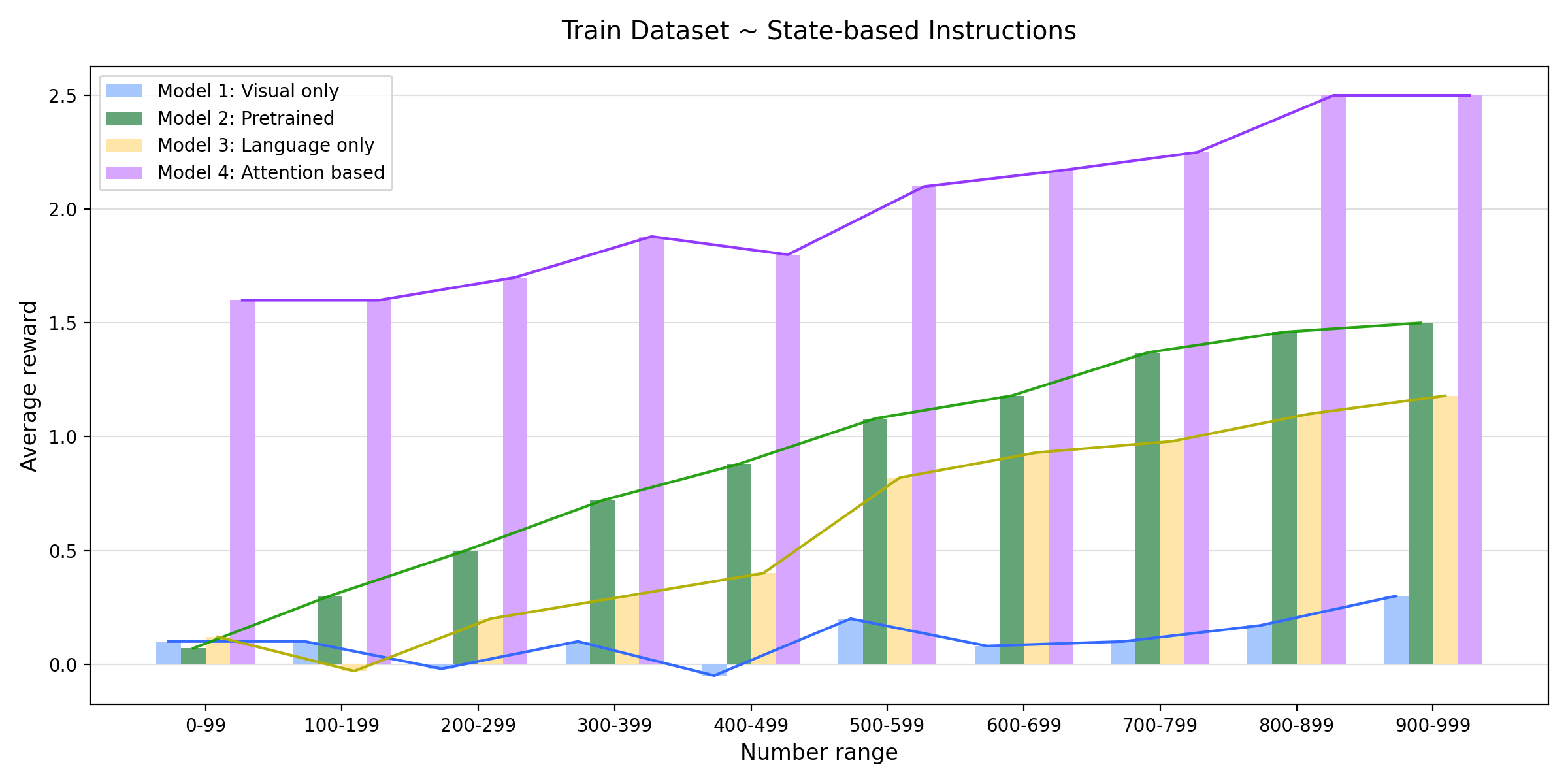}
	\caption{Average reward for training data across different number groupings when state-based instructions were used to guide the RL agent.}
	\label{StateTrain}
\end{figure}

\begin{figure}[!htbp]
	\centering
	\includegraphics[width=0.5\textwidth]{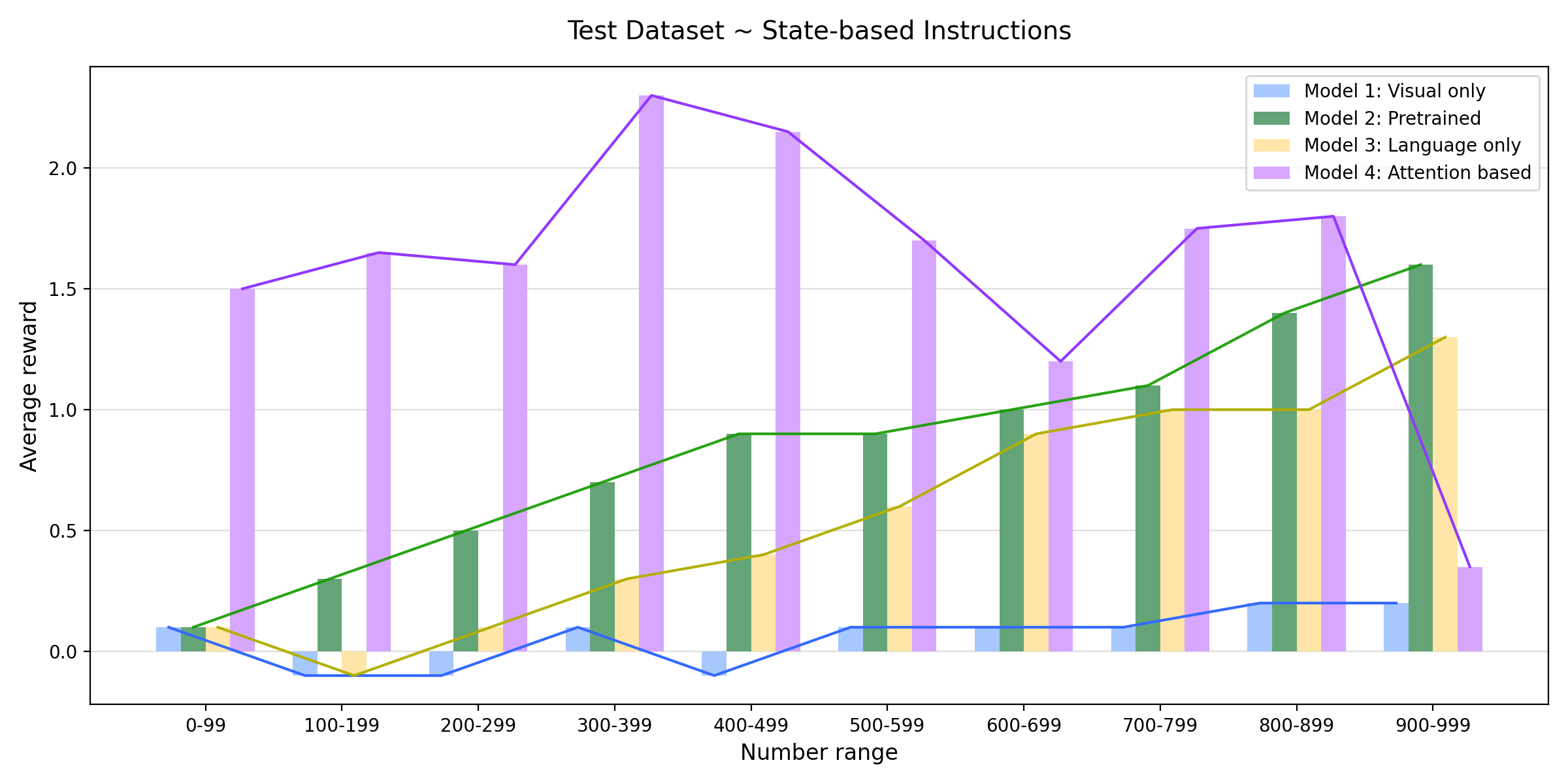}
	\caption{Average reward for test dataset across different number groupings when state-based instructions were used to guide the RL agent.}
	\label{StateTest}
\end{figure}

Figure \ref{StateTrain} and \ref{StateTest} shows the average reward for the training and test datasets respectively for state-based instructions. Attention-based neural net i.e. "Model 4" was the only model to perform relatively well on the task.

\begin{figure}[!htbp]
	\centering
	\includegraphics[width=0.5\textwidth]{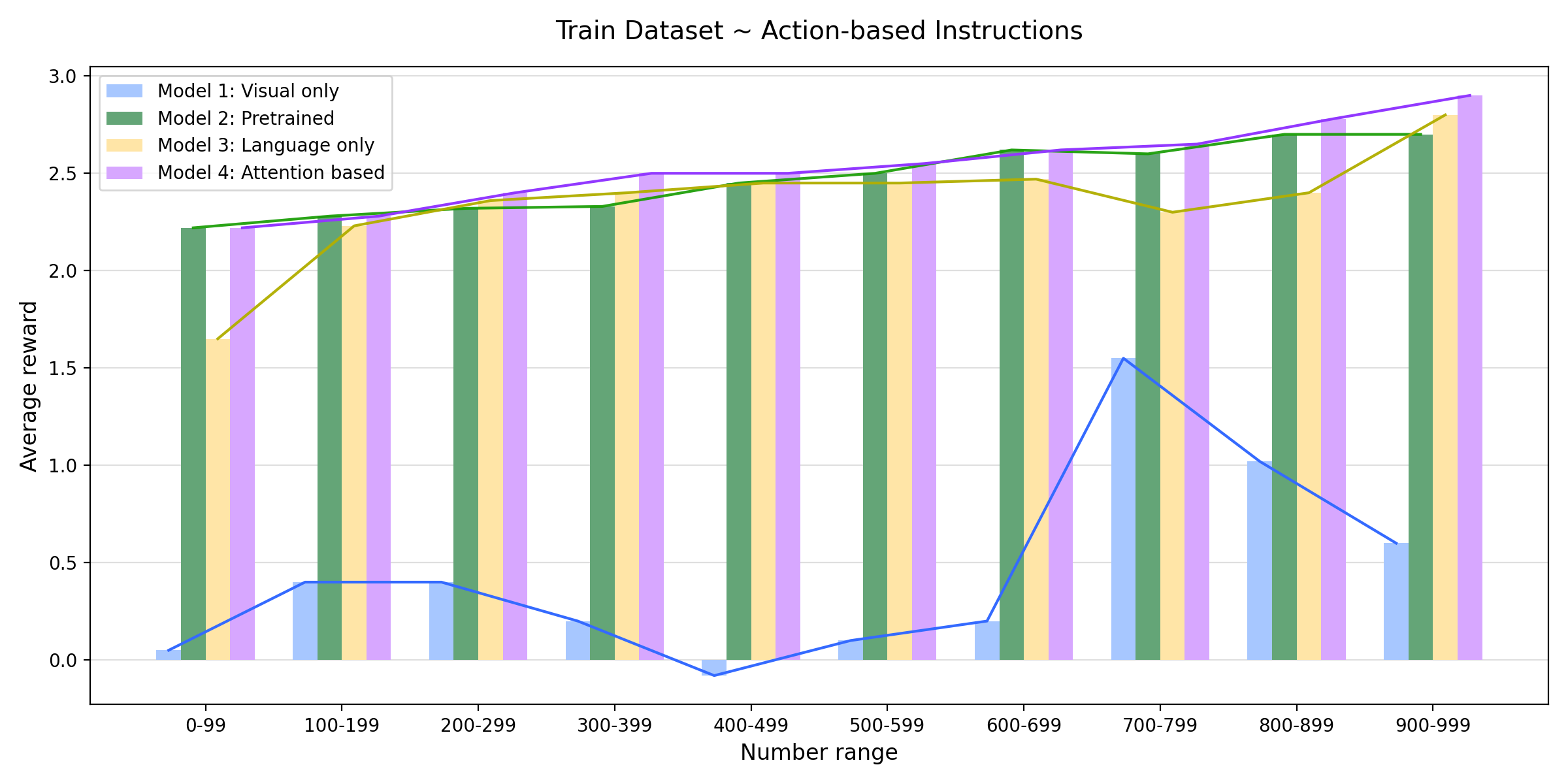}
	\caption{Average reward for training data across different number groupings when action-based instructions were used to guide the RL agent.}
	\label{ActionTrain}
\end{figure}

\begin{figure}[!htbp]
	\centering
	\includegraphics[width=0.5\textwidth]{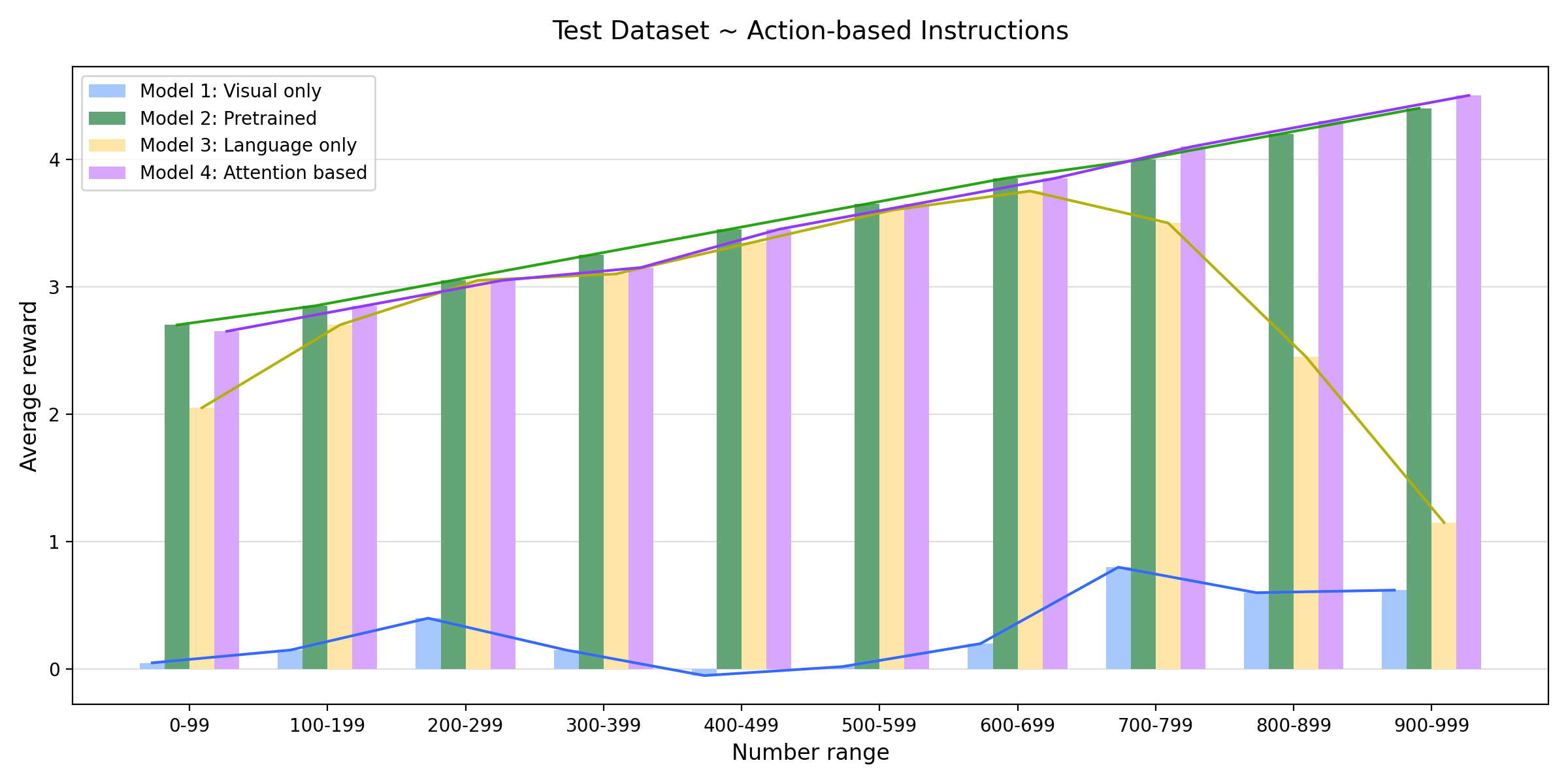}
	\caption{Average reward for test dataset across different number groupings when action-based instructions were used to guide the RL agent.}
	\label{ActionTest}
\end{figure}

Figure \ref{ActionTrain} and \ref{ActionTest} shows the average reward for the training and test datasets respectively for action-based instructions. Agents that ignored the language input and focused only on visual cues performed worse and were not able to generalize to unknown numbers.

\end{document}